\documentclass[conference]{IEEEtran}
\usepackage{times}

\usepackage[numbers]{natbib}
\usepackage{multicol}
\usepackage[bookmarks=true]{hyperref}

\usepackage{textcomp}
\usepackage{wrapfig}
\usepackage{graphicx}
\usepackage{caption}
\usepackage{float}
\usepackage{amsmath}
\usepackage{amsfonts}
\usepackage{amssymb}
\usepackage{colortbl}  
\usepackage{xcolor}  
\usepackage{algpseudocode}
\usepackage{algorithmicx}
\usepackage{algorithm}
\usepackage{makecell}  
\usepackage{multirow}  
\usepackage{pifont}

\let\oldtwocolumn\twocolumn
\renewcommand\twocolumn[1][]{
    \oldtwocolumn[{#1}{
	\begin{center}
           \includegraphics[width=0.95\textwidth]{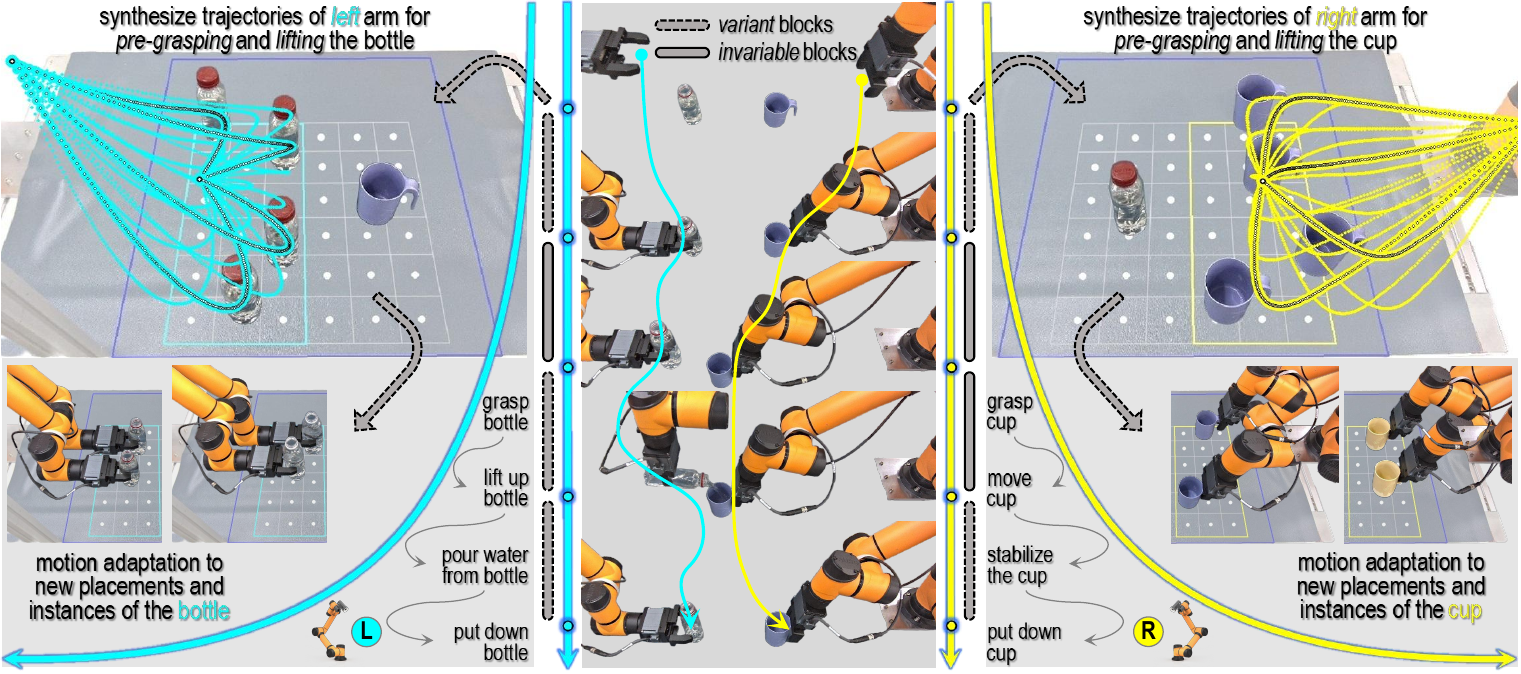}
	\vspace{-5pt}
	\captionof{figure}{\textbf{From One to Many \textcircled{\scriptsize 1}$\rightarrow$\textcircled{\scriptsize N}}. Taking the dual-arm coordinated \texttt{pouring} water task as an example, we illustrate how to synthesize corresponding pre-grasping and lifting trajectories for different new placements and novel instances of manipulated objects (\textit{e.g.}, bottle and cup) during the initial frame alignment phase for pre-grasping.} 
           \label{BiDemoSyn}
	\end{center}
    }]
}

\begin{document}

\title{One-Shot Real-World Demonstration Synthesis for Scalable Bimanual Manipulation}


\author{
\authorblockN{
Huayi Zhou\authorrefmark{1} and
Kui Jia\authorrefmark{1}\authorrefmark{2}\authorrefmark{3}
}
\authorblockA{
\authorrefmark{1}The Chinese University of Hong Kong, Shenzhen.\quad
\authorrefmark{2}DexForce, Shenzhen.\quad
\authorrefmark{3}The Corresponding Author.}
\url{https://hnuzhy.github.io/projects/BiDemoSyn}
\vspace{-15pt}
}

\maketitle
\begin{abstract}
Learning dexterous bimanual manipulation policies critically depends on large-scale, high-quality demonstrations, yet current paradigms face inherent trade-offs: teleoperation provides physically grounded data but is prohibitively labor-intensive, while simulation-based synthesis scales efficiently but suffers from sim-to-real gaps. We present BiDemoSyn, a framework that synthesizes contact-rich, physically feasible bimanual demonstrations from a single real-world example. The key idea is to decompose tasks into invariant coordination blocks and variable, object-dependent adjustments, then adapt them through vision-guided alignment and lightweight trajectory optimization. This enables the generation of thousands of diverse and feasible demonstrations within several hours, without repeated teleoperation or reliance on imperfect simulation. Across six dual-arm tasks, we show that policies trained on BiDemoSyn data generalize robustly to novel object poses and shapes, significantly outperforming recent strong baselines. Beyond the one-shot setting, BiDemoSyn naturally extends to few-shot-based synthesis, improving object-level diversity and out-of-distribution generalization while maintaining strong data efficiency. Moreover, policies trained on BiDemoSyn data exhibit zero-shot cross-embodiment transfer to new robotic platforms, enabled by object-centric observations and a simplified 6-DoF end-effector action representation that decouples policies from embodiment-specific dynamics. By bridging the gap between efficiency and real-world fidelity, BiDemoSyn provides a scalable path toward practical imitation learning for complex bimanual manipulation without compromising physical grounding.
\end{abstract}

\IEEEpeerreviewmaketitle
\section{Introduction}

Recent advances in robot manipulation have been propelled by data-driven imitation learning, where visuomotor policies trained on large-scale demonstration datasets enable dexterous single-arm and bimanual tasks previously deemed intractable. Methods like ACT \cite{zhao2023learning}, DP \cite{chi2023diffusion} and $\pi_0$ \cite{black2024pi0} exemplify this trend, achieving remarkable performance in contact-rich scenarios such as continuously pushing, battery assembly, and coordinated garment folding. These successes, however, hinge on a critical yet underappreciated premise: the availability of diverse, high-quality demonstrations, which are typically collected via expert teleoperation. As the community shifts toward more complex long-horizon bimanual tasks, the data scalability bottleneck becomes increasingly apparent. Even those largest manipulation datasets like RH20T \cite{fang2024rh20t}, DROID \cite{khazatsky2024droid}, Open X-Embodiment \cite{o2024open} and AgiBot World \cite{bu2025agibot} struggle to provide sufficient geometric and kinematic diversity for robust generalization, underscoring a pivotal question: \textit{how to scale real-world bimanual demonstration collection without compromising physical fidelity?}

The answer, unfortunately, lies in a fundamental trade-off. Existing data acquisition methods fall into two paradigms, each with critical shortcomings. On one hand, human-operated systems (\textit{e.g.}, the ALOHA series \cite{zhao2023learning, fu2024mobile, zhao2024aloha}) yield physically-accurate trajectories but demand prohibitive expertise, rendering them impractical for scaling to long-horizon tasks. On the other hand, simulation-based methods (\textit{e.g.}, MimicGen \cite{mandlekar2023mimicgen, jiang2024dexmimicgen}, RoboGen \cite{wang2024robogen}, and RoboCasa \cite{nasiriany2024robocasa}) bypass human labor through programmatic data generation utilizing massive 3D assets, yet discrepancies in contact dynamics and visual rendering inevitably plague real-world deployment \cite{bharadhwaj2024position, mccarthy2024towards}. Thus, the tension between data quality (grounded in reality) and quantity (enabled by automation) persists as a key barrier to scalable imitation learning, particularly for contact-rich bimanual tasks requiring precise dual-arm coordination.

To bridge this gap, we present \textbf{BiDemoSyn} that synthesizes real-world bimanual demonstrations from a single exemplar, by algorithmically amplifying task semantics through a hierarchical decomposition of invariance and adaptability (refer Fig.~\ref{BiDemoSyn} and Fig.~\ref{framework}). Our motivation is inherited from the well-established single-arm one-shot imitation learning \cite{mao2023learning, liu2025one, biza2023one, zhang2023universal, dreczkowski2025learning, zhou2025vlbiman}. The key insight lies in recognizing that complex manipulation tasks exhibit a duality: while certain motion primitives (\textit{e.g.}, stable grasp sequences, dual-arm synchronization) remain invariant across instances, others (\textit{e.g.}, trajectory segments adapting to object shifts) require dynamic adjustments to contextual variations. Specifically, through a three-stage process, \textbf{BiDemoSyn} first deconstructs the initial demonstration into invariant primitives and geometry-sensitive components. Then, it leverages visual perception to spatially align these elements in novel scenes, enabling generalization across object geometries and workspace layouts with minor effort (primarily involving reinitializing objects). Finally, a physics-aware optimizer jointly modulates dual-arm trajectories, injecting diversity into adaptable components while enforcing real-world constraints on collision avoidance and kinematic coordination. This integrated approach achieves \textit{One-to-Many} demonstration synthesis: a single exemplar spawns hundreds of trajectories that preserve task intent while adapting to unseen configurations, all grounded in real-world dynamics rather than simulated proxies.

Unlike prior simulation-based methods \cite{mandlekar2023mimicgen, wang2024robogen, nasiriany2024robocasa, jiang2024dexmimicgen}, \textbf{BiDemoSyn} operates entirely in the physical domain, ensuring synthesized data inherits the fidelity of human demonstrations. By unifying algorithmic scalability with physical realism, our framework empowers policies to transcend the limitations of narrow manually-collected data distributions. Experiments on real dual-arm platforms validate the effectiveness of \textbf{BiDemoSyn}. Visuomotor policies trained with synthesized demonstrations can achieve superior success rates and generalization in tasks requiring precise contact coordination (\textit{e.g.}, small hole insertion, rotation angle control, purposeful handover and articulated object manipulation).

Our contributions are threefold: (1) \textit{One-Shot Synthesis Framework}: A systematic pipeline combining task decomposition, vision-guided adaptation, and contact-aware trajectory optimization to generate scalable real-world bimanual demonstrations. (2) \textit{Reality-Grounded Data Generation}: A completely simulator-free method for synthesizing bimanual demonstrations, ensuring physical fidelity by construction. (3) \textit{Empirical Validation in Complex Tasks}: Comprehensive real-robot experiments demonstrating significant improvements in policy robustness and cross-configuration generalization on various bimanual manipulation tasks.

\section{Related Works}

\textbf{Bimanual Robotic Manipulation.}  
Prior literatures primarily target task-specific challenges (such as cloth folding \cite{colome2018dimensionality, weng2022fabricflownet, canberk2023cloth}, untangling \cite{grannen2021untangling, peng2024tiebot}, untwisting \cite{lin2024twisting, bahety2024screwmimic}, and handover \cite{huang2023dynamic, li2023efficient}) through handcrafted controllers or narrow skill libraries, limiting generalization due to rigid motion primitives and excessive assumptions. General methods \cite{xie2020deep, chitnis2020efficient, chen2022towards, chen2023bi, hartmann2022long, zhaodual2023afford, zhou2025vlbiman} either rely on predefined hierarchical arm roles (\textit{e.g.}, leader-follower \cite{krebs2022bimanual, liu2022robot, grotz2024peract2} and stabilizer-actor \cite{grannen2023stabilize, liu2024voxact}) or hinge generalization entirely on dataset diversity, both of which are ultimately bottlenecked by the impractical costs of data acquisition. Recently, ALOHA series \cite{zhao2023learning, fu2024mobile, zhao2024aloha} have revolutionized bimanual manipulation by dexterous low-cost teleoperation. Meanwhile, the visuomotor policy learning paradigm \cite{chi2023diffusion, ze2024dp3, yang2024equibot} based on diffusion models \cite{ho2020denoising, song2021denoising} has largely expanded the action modeling space and data utilization efficiency. Their followers \cite{team2024octo, kim2024openvla, liu2025rdt, black2024pi0, pertsch2025fast, lin2025data} expect to train universal policies using large-scale teleoperation data but face scalability barriers from human effort. To improve reachability and dexterity, some studies use specialized hardware (\textit{e.g.}, multi-finger hands \cite{wang2024dexcap, shaw2024bimanual, fu2024humanplus, cheng2024open} or tactile sensors \cite{lin2024learning, chen2024arcap}), which complicate real-world adoption. In contrast, we utilize a fixed-base dual-arm platforrm with parallel grippers, synthesizing diverse demonstrations from a single example. By optimizing coordination as dynamic constraints rather than predefined hierarchies, we bypass hardware specificity and data scalability limitations, enabling task-agnostic policies grounded in physical feasibility.

\textbf{Bimanual Demonstration Collection.}  
The dominant path for acquiring bimanual demonstrations is \textit{human teleoperation} \cite{khazatsky2024droid, bu2025agibot}, which delivers high-quality data but suffers from prohibitive scalability costs. To alleviate it, two alternative routes have emerged: \textit{simulation-based synthesis} \cite{garrett2024skillmimicgen, hua2024gensim2, liang2024make, yang2024physcene, wang2024gensim} and \textit{learning from human videos} \cite{ponimatkin20256d, ye2025video2policy, zhao2025taste, kareer2024egomimic, bharadhwaj2024track2act}. The former, exemplified by MimicGen \cite{mandlekar2023mimicgen} and DexMimicGen \cite{jiang2024dexmimicgen}, augments a few demonstrations by generating variations in simulation using geometric transformations and kinematic constraints. Analogously, RoboGen \cite{wang2024robogen} and RoboCasa \cite{nasiriany2024robocasa} leverage 3D assets to procedurally generate full-scene manipulation data, yet such methods inevitably inherit \textit{sim-to-real} gaps, from unrealistic textures to unphysical dynamics. The latter extracts bimanual hand-object manipulation from egocentric videos \cite{zhan2024oakink2, liu2024taco, grauman2024ego} and maps them to dual arms via motion retargeting \cite{li2024okami, kerr2024robot, chen2024object, chen2024vividex} or non-privileged representations (\textit{e.g.}, keypoints \cite{papagiannis2024rx, gao2024bi, wen2023any}, affordances \cite{ju2024robo, nasiriany2024rt} and correspondences \cite{peng2024tiebot, ko2024learning, zhang2024one}). While human videos are abundant and low-cost, large morphological mismatches between humanity and robotic embodiments often need heuristic translation rules and hinder direct applicability. Our work navigates these trade-offs by synthesizing demonstrations directly in real-world. Given a single exemplar, we diversify trajectories through vision-based adaptation and physics-compliant optimization. This balances the fidelity and scalability, while avoiding retargeting hurdles, thereby enabling practical and scalable data acquisition for contact-rich bimanual tasks.

Several concurrent works exhibit notable limitations. DemoGen \cite{xue2025demogen} and YOTO \cite{zhou2025you, zhou2026yoto++} diversify demos via 3D editing of the initial single-view point cloud, but perspective ambiguities induce visual artifacts in synthesized data. ODIL \cite{wang2025one} avoids editing by segmenting objects for visual servoing and iteratively replanning trajectories, which is cumbersome for scalability. MoMaGen \cite{li2025momagen} is a long-horizon mobile bimanual demonstrations generator yet mainly focusing on the simulation. Our \textbf{BiDemoSyn} sidesteps these issues: synthesizing trajectories directly in real-world ensures visual-physical consistency without manual edits or robot replaying.

\section{Preliminaries}\label{SecPre}

\textbf{Problem Formulation.} 
Bimanual imitation learning aims to train visuomotor policies $\pi_\theta(\textbf{o}_t) \rightarrow \textbf{a}_t$, where $\textbf{o}_t$ denotes multimodal observations (\textit{e.g.}, RGB images, joint states or point clouds) and $\textbf{a}_t$ represents dual-arm actions. While architectures like ACT \cite{zhao2023learning} and DP \cite{chi2023diffusion} enhance policy expressivity, their generalization critically depends on expert demonstrations $\mathcal{D}\!=\!\{ \tau_i \}_{i=0}^N$ that densely span the feasible state-action manifold. Existing approaches either collect $\mathcal{D}$ via labor-intensive teleoperation or generate data in simulation, both failing to balance real-world fidelity and scalability. We address this gap by proposing the following problem:

Consider a bimanual task defined by an initial state $s_0$ and a specific goal $g$. Given a single real-world demonstration $\tau\!=\!\{ \textbf{o}_t, \textbf{a}_t \}_{t=0}^T$, our objective is to synthesize a dataset $\mathcal{D}_{syn}\!=\!\{ \tau_i \}_{i=0}^M (M \gg 1)$ such that: (1) \textit{Task Consistency}: Every $\tau_i \in \mathcal{D}_{syn}$ achieves the task goal $g$ under perturbed initial states (\textit{e.g.}, new object poses, scene layouts). (2) \textit{Physical Admissibility}: Trajectories adhere to kinematic and dynamic constraints of the real-world system (\textit{e.g.}, collision avoidance, dual-arm coordination). (3) \textit{Diversity}: $\mathcal{D}_{syn}$ covers task-relevant variations to enable robust policy training.

\textbf{Visuomotor Policy Learning.}
Our $\pi_\theta$ maps observations $\textbf{o}_t$ to dual-arm actions $\textbf{a}_t\!=\!\left[ \textbf{a}_t^L, \textbf{a}_t^R \right]\!\in\!\mathbb{R}^{2d}$, where $\textbf{a}_t^L$, $\textbf{a}_t^R$ denotes left/right end-effector poses. We adapt a diffusion policy model \cite{chi2023diffusion, ze2024dp3, yang2024equibot} to generate action sequences $\textbf{A}\!=\!\left[ \textbf{a}_{t:{t+H}}^L, \textbf{a}_{t:{t+H}}^R \right]\!\in\!\mathbb{R}^{2H \times d}$ over a horizon $H$. The diffusion process iteratively denoises a noisy action sequence $\textbf{A}_k$ toward kinematic feasibility over $K$ steps:
\begin{equation}
	\centering
	\textbf{A}_k = \sqrt{\alpha_k} \textbf{A}_0 + \sqrt{ 1 - \alpha_k} \epsilon, \quad \epsilon \sim \mathcal{N}(0, 
\mathbf{\Sigma}),
	\label{denoiseEqn}
\end{equation}
where $\alpha_k$ follows a cosine schedule \cite{nichol2021improved}. And $\mathbf{\Sigma}\!\in\!\mathbb{R}^{2H \times 2H}$ is a block-diagonal covariance matrix {\scriptsize $\left[\begin{array}{cc} \mathbf{\Sigma}_L&\rho\mathbf{\Sigma}_{LR}\\\rho\mathbf{\Sigma}^{\top}_{LR}&\mathbf{\Sigma}_R \end{array}\right]$} encoding arm coordination priors, where $\mathbf{\Sigma}_L$, $\mathbf{\Sigma}_R$ governing temporal smoothness per arm, $\mathbf{\Sigma}_{LR}$ modeling inter-arm dependencies, and $\rho\!\in\![0,1]$ controlling coordination strength. Then, a denoiser $\mu_\theta$ processes the noised sequence conditioned on visual observations $\textbf{o}_t$ via:
\begin{equation}
	\centering
	\mu_\theta (\textbf{A}_k, \textbf{o}_t, k) = \texttt{UNet}_\theta (\texttt{Concat}\left[ \textbf{A}_k, \mathsf{E}(\textbf{o}_t) \right], k),
	\label{dpEqn}
\end{equation}
where $\mathsf{E}(\textbf{o}_t) $ is a vision encoder (ResNet \cite{he2016deep} or PointNet \cite{qi2017pointnet}) extracting latent features. The $\texttt{UNet}_\theta$ employs asymmetric layers with early stages processing each arm independently, while deeper layers fusing bilateral coordinated features. Training minimizes:
\begin{equation}
	\centering
	\mathcal{L}(\theta) = \mathbb{E}_{\textbf{A}_0, \textbf{o}_t, k}\left[ \left\Vert \mu_\theta (\textbf{A}_k, \textbf{o}_t, k) - \textbf{A}_0 \right\Vert_{\textbf{W}}^2 \right],
	\label{lossEqn}
\end{equation}
where $\|\!\cdot\!\|_{\textbf{W}}$ is a Mahalanobis norm with $\textbf{W}\!=\!\mathbf{\Sigma}^{-1}$ to prioritize arm interactions kinematically. By training on $\mathcal{D}_{syn}$ with feasible trajectories synthesized from a single exemplar, the policy bypasses sim-to-real gaps while handling diverse bimanual coordination patterns.

\begin{figure*}
	\begin{center}
           \includegraphics[width=1.0\linewidth]{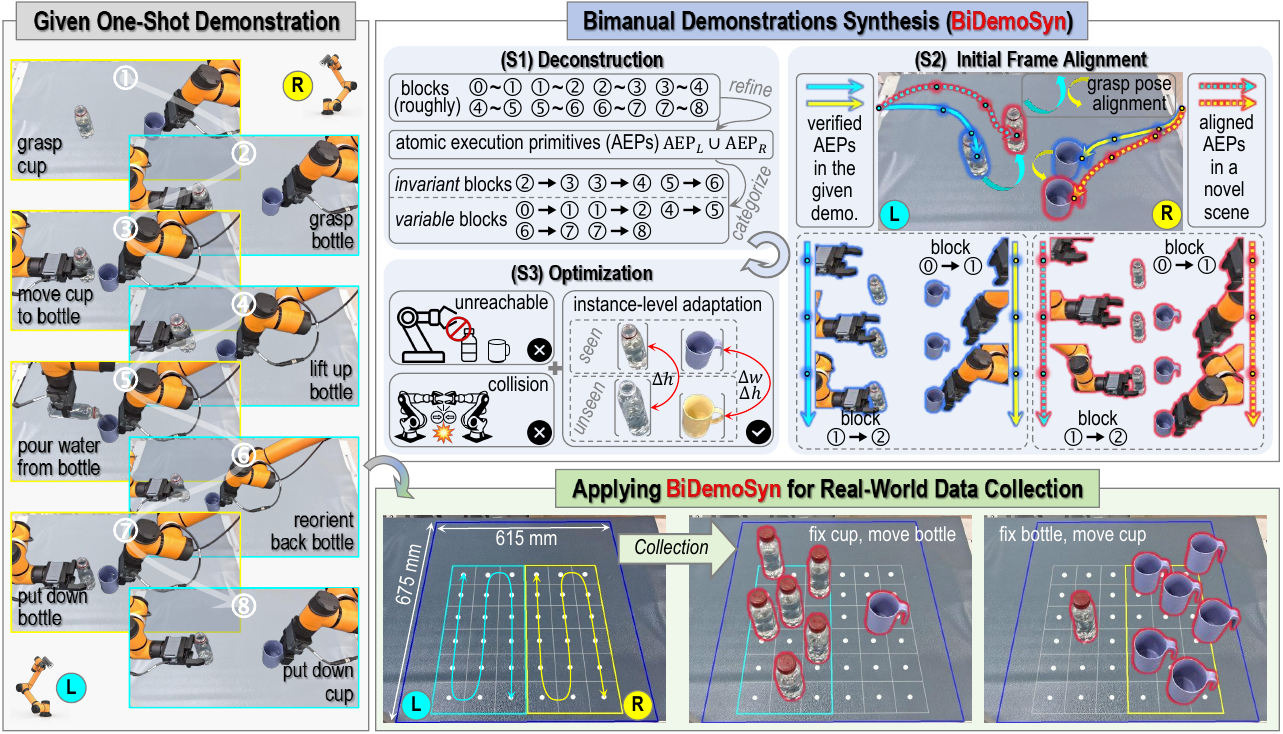}
	\vspace{-18pt}
	\caption{The overview of \textbf{BiDemoSyn}. It consists of three stages (\textit{e.g.}, deconstruction, alignment, and optimization) based on a given demonstration. Then, we can apply our method to complete data collection efficiently and conveniently in real-world.} 
           \label{framework}
	\vspace{-20pt}
	\end{center}
\end{figure*}

\section{Methodology}\label{method}
Here, we detail \textbf{BiDemoSyn} for synthesizing bimanual demonstrations $\mathcal{D}_{syn}$ from a single exemplar $\tau$ with three compact stages (see Fig.~\ref{framework}): Sec.~\ref{DOST} \textit{Deconstruction of One-Shot Teaching}, which extracts invariant patterns and adaptable primitives; Sec.~\ref{VIFA} \textit{Vision-based Initial Frame Alignment}, enabling generalization across geometric variations via efficient scene perception; and Sec.~\ref{TMO} \textit{Trajectory Modulation and Optimization}, ensuring physical feasibility via hierarchical kinematic constraints. Each stage sequentially addresses scalability, adaptability, and real-world fidelity to bridge the gap between data efficiency and policy robustness.

\subsection{Deconstruction of One-Shot Teaching}\label{DOST}
Given a demonstration $\tau\!=\!\{ \textbf{o}_t, \textbf{a}_t \}_{t=0}^T$, we decompose it into a \textit{bimanual execution blocks} set $\{ \mathcal{B}_i \}_{i=1}^n$, where each block $\mathcal{B}_i\!=\!(\textbf{s}_i^L, \textbf{s}_i^R)$ represents a discrete phase of dual-arm interaction. Here, $\textbf{s}_i^L, \textbf{s}_i^R\!\in\!\mathbb{R}^d$ denote the left/right arm state sequences (\textit{e.g.}, end-effector 6-DoF poses and gripper status) within block $\mathcal{B}_i$. Blocks are categorized based on motion coordination. For \textbf{single-arm motion}, one arm exhibits significant transitions while the other remains static. Formally, for threshold $\delta$ (minimal motion saliency) and $\zeta$ (static tolerance):
\begin{equation}
\setlength{\jot}{1pt}
\begin{aligned}
	\centering
	&(\left\Vert \textbf{s}_{i,e}^L - \textbf{s}_{i,s}^L \right\Vert\!\geq\!\delta \wedge \left\Vert \textbf{s}_{i,e}^R - \textbf{s}_{i,s}^R \right\Vert\!\leq\!\zeta) \; \vee \; \\
	&(\left\Vert \textbf{s}_{i,e}^R - \textbf{s}_{i,s}^R \right\Vert\!\geq\!\delta \wedge \left\Vert \textbf{s}_{i,e}^L - \textbf{s}_{i,s}^L \right\Vert\!\leq\!\zeta).
	\label{singlearmEqn}
\end{aligned}
\end{equation}
While, for \textbf{dual-arm coordination}, both arms synchronously or asynchronously adjust states:
\begin{equation}
	\centering
	\left\Vert \textbf{s}_{i,e}^L - \textbf{s}_{i,s}^L \right\Vert\!\geq\!\delta \wedge \left\Vert \textbf{s}_{i,e}^R - \textbf{s}_{i,s}^R \right\Vert\!\geq\!\delta.
	\label{dualarmEqn}
\end{equation}

\subsubsection{From Blocks to Atomic Execution Primitives (AEPs)}
To enable modular adaptation, each block $\mathcal{B}_i$ is further refined into a minimal motion unit AEP, where an arm undergoes a salient state transition (\textit{e.g.}, gripper closure, moving or pose shifts). For arm $A\!\in\!\{L,R\}$, an AEP is defined as:
\begin{equation}
\setlength{\jot}{1pt}
\begin{aligned}
	\centering
	\texttt{AEP}_{A} = &\{ \textbf{s}_t^A \rightarrow \textbf{s}_{t+\Delta t}^A \; | \; \left\Vert \textbf{s}_{t+\Delta t}^A - \textbf{s}_t^A \right\Vert_2 \geq \gamma, \\ 
	& \Delta t \leq T_\textrm{\scriptsize max}, \texttt{CollisionFree}(\textbf{s}_{[t, \Delta t]}^A) \},
	\label{aepEqn}
\end{aligned}
\end{equation}
where $\gamma$ is a distance threshold, $T_\textrm{\scriptsize max}$ limits execution duration, and $\texttt{CollisionFree}(\cdot)$ enforces no collisions between the arm and objects/scene (optimized via forward kinematics). Unlike keyframe-based segments \cite{james2022coarse, shridhar2023perceiver, ma2024hierarchical, zhou2025you}, AEPs ignore acceleration profiles (assuming quasi-static motions) and prioritize task-oriented transitions over temporal granularity.

\subsubsection{Semantic Categorization for Adaptive Synthesis}
The final step categorizes refined blocks into \textit{invariant} and \textit{variable} types. The invariant blocks encode task-semantic primitives (\textit{e.g.}, screwing, pressing) or general motions (\textit{e.g.}, lifting, transferring), which remain structurally consistent across variations. The variable blocks, such as object-centric goal-conditioned grasping, adapt to instance-level geometric variations including object poses and shapes through:
\begin{equation}
	\centering
	\mathcal{B}'_i= \mathcal{B}_i \circ \Phi(g, \textbf{o}_\textrm{\scriptsize novel}),
	\label{visadaEqn}
\end{equation}
where $\Phi(\cdot)$ is a visual adapter, $g$ is the task goal, and $\textbf{o}_\textrm{\scriptsize novel}$ is the novel observation. This structured decomposition enables selective diversification, allowing variable blocks to adapt to new scenes while preserving task fidelity in invariant blocks.

\subsection{Vision-based Initial Frame Alignment}\label{VIFA}
\begin{figure*}
	\begin{center}
           \includegraphics[width=\linewidth]{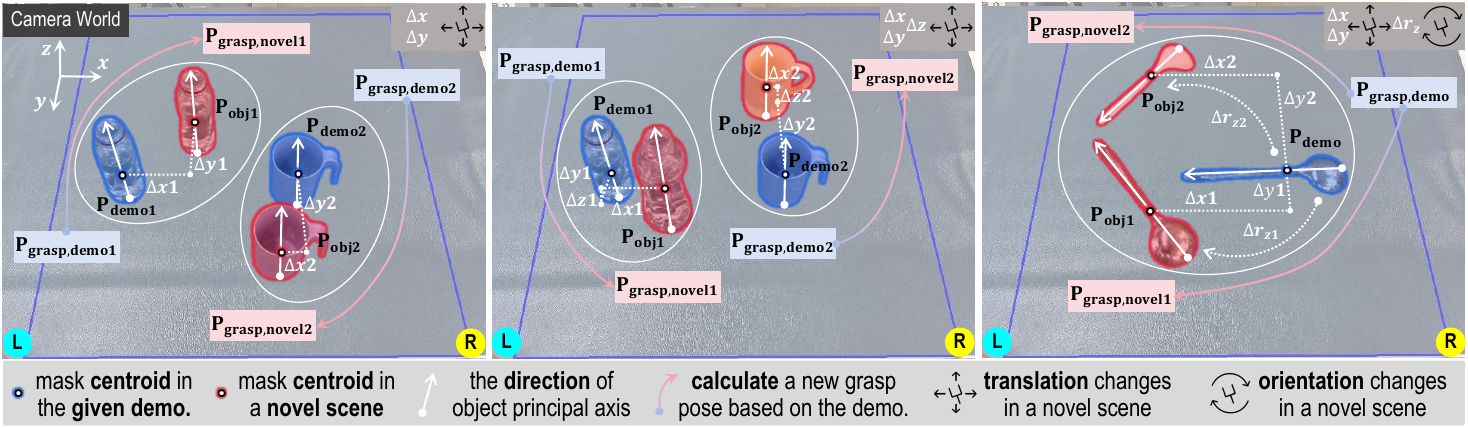}
	\vspace{-15pt}
	\caption{Illustrations of the initial frame alignment applied to tasks \texttt{pouring} (left and middle) and \texttt{reorient} (right). It shows that we can automatically adjust the grasp pose after the position, orientation and shape of the manipulated object changes.}
           \label{alignment}
	\vspace{-20pt}
	\end{center}
\end{figure*}

The variable blocks identified in Sec.~\ref{DOST} require adapting to novel object poses and geometries. To achieve this, we propose a vision-driven adapter $\Phi(\cdot)$ that generalizes object-centric interactions in the given exemplar to new scenes. Given a novel observation $\textbf{o}_\textrm{\scriptsize novel}$, our method executes three steps: \textit{object perception}, \textit{state estimation}, and \textit{pose alignment}, ensuring the initial robotic grasp aligns with the task goal $g$. Illustrations are shown in Fig.~\ref{alignment}.

\subsubsection{Object Perception}
We first detect and segment the target object in $\textbf{o}_\textrm{\scriptsize novel}$ using an open-vocabulary detector (\textit{e.g.}, YOLO-World \cite{cheng2024yolo} or YOLOE \cite{wang2025yoloe}) or vision foundation models (\textit{e.g.}, Florence2 \cite{xiao2024florence} with SAM2 \cite{ravi2024sam} for robustness to rare categories). In practice, we prioritize foundation models to avoid detection failures. Now let $M\!\subseteq\!\textbf{o}_\textrm{\scriptsize novel}$ denote the obtained object's binary mask.

\subsubsection{State Estimation}
Instead of uitlizing category-level CAD-based 6D pose estimation models (\textit{e.g.}, FoundationPose \cite{wen2024foundationpose}), we estimate the instance-level 6D object pose $\mathbf{P}_\textrm{\scriptsize obj}\!\in\!SE(3)$ using geometry-aware processing. Sepcifically, we adopt classic image moments \cite{chaumette2004image, kotoulas2007accurate} to calculate the object mask centroid $\textbf{c}$ and extract principal axes $\textbf{R}$:
\begin{equation}
	\centering
	\left\{ \begin{array}{ccl} \textbf{c} &=& \frac{1}{|M|}\sum_{(u,v) \in M}(u,v,\textrm{d}(u,v)) \\ \textbf{R} &=& \texttt{PCA}(\{ (u,v,\textrm{d}(u,v)) \; | \; (u,v) \in M \}) \end{array}\right.
	\label{imgmomEqn}
\end{equation}
where $\textrm{d}(u,v)$ is the depth value. $\texttt{PCA}$ is used to fit 3D points of $M$ to determine orientation $\textbf{R}\!\in\!SO(3)$. This yields $\mathbf{P}_\textrm{\scriptsize obj}\!=\!(\textbf{R}, \textbf{c})$, robust to arbitrary object states (e.g., fallen, inverted).

\subsubsection{Pose Alignment}
Let $\mathbf{P}_\textrm{\scriptsize demo}$ denote the object pose in the given demonstration. We compute the rigid transformation $\mathbf{T}\!\in\!SE(3)$ that maps $\mathbf{P}_\textrm{\scriptsize demo}$ to $\mathbf{P}_\textrm{\scriptsize obj}$. This transformation is applied to the initial grasp pose $\mathbf{P}_\textrm{\scriptsize grasp,demo}$ in the variable block $\mathcal{B}_i$, yielding the adapted grasp pose:
\begin{equation}
	\centering
	\mathbf{P}_\textrm{\scriptsize grasp,novel} = \mathbf{T} \cdot \mathbf{P}_\textrm{\scriptsize grasp,demo} = (\mathbf{P}_\textrm{\scriptsize obj} \cdot \mathbf{P}^{-1}_\textrm{\scriptsize demo}) \cdot \mathbf{P}_\textrm{\scriptsize grasp,demo}.
	\label{alignEqn}
\end{equation}
The robot arm then executes a collision-free motion trajectory to reach $\mathbf{P}_\textrm{\scriptsize grasp,novel}$, ensuring goal-conditioned grasping without relying on task-irrelevant dense grasp proposals produced via 6-DoF grasp pose detectors \cite{fang2020graspnet, fang2023anygrasp}. By integrating with the proposed decomposition stage, the adapted grasp pose directly modifies the variable block $\mathcal{B}_i$ into $\mathcal{B}'_i$, enabling synthesis of new trajectories.

\subsection{Trajectory Modulation and Optimization}\label{TMO}
After deconstructing the provided demonstration into blocks (Sec.~\ref{DOST}) and adapting variable blocks via vision-based alignment (Sec.~\ref{VIFA}), the synthesized trajectory still requires two critical refinements to ensure executability and robustness as expalined below:

\subsubsection{Collision-Aware Validation}
Each adapted block undergoes kinematic feasibility checks. First, we validate the reachability of target poses $\mathbf{P}_\textrm{\scriptsize grasp,novel}$ using Inverse Kinematics (IK), ensuring the robot joint limits are satisfied. If the result returned is singular or unreachable (rarely happens), we will delete this sample. Second, a motion planner \cite{chitta2012moveit, schulman2014motion} verifies potential collision between arms and scene objects by solving for collision-free paths between the start and end states in each block. This simplified two-point constraint reduces computational overhead while preserving safety, assuming the trajectory of original given demonstration is collision-free.

\subsubsection{Instance-Level Motion Adaptation}
To handle geometric variations across object instances with differing shapes, we adjust motion primitives in variable blocks based on the object 3D bounding box dimensions. Let $\Delta l$, $\Delta w$ and $\Delta h$ denote the length, width and height differences between the novel object bounding box $\textbf{b}_\textrm{\scriptsize novel}$ and the original $\textbf{b}_\textrm{\scriptsize demo}$. Motion endpoints (\textit{e.g.}, grasp or release positions) are offset by:
\begin{equation}
	\centering
	\mathbf{s}_{i,e}^{A'} = \mathbf{s}_{i,e}^{A} + \lambda(\Delta l, \Delta w, \Delta h),
	\label{offsetEqn}
\end{equation}
where $\lambda$ scales the adjustment (empirically set to $0.8\sim1.0$), and $A\!\in\!\{L,R\}$. For irregular shapes, depth-aware masks $M$ extracted in Sec.~\ref{VIFA} can approximate volumetric differences, or minimal human input provides precise measurements.

This hierarchical optimization stage operates at the block level rather than full trajectories, and enables efficient scaling while preserving the task semantics encoded in invariant blocks. Finally, validated and adapted blocks are recombined into synthetic trajectories $\tau'\!=\!\{ \tau_\textrm{\scriptsize inv}, \{\mathcal{B}'_i\} \}$, populating diverse and physically feasible demonstrations $\mathcal{D}_{syn}$.

\section{Experiments}
Our experiments address five core questions. $\mathsf{Q}1$: Is BiDemoSyn truly efficient and user-friendly? $\mathsf{Q}2$: Does BiDemoSyn enable scalable visuomotor imitation learning? $\mathsf{Q}3$: Does synthetic demonstrations generalize to spatial and object variations? $\mathsf{Q}4$: Can BiDemoSyn seamlessly integrate with few-shot demonstrations to further improve robustness and data efficiency? $\mathsf{Q}5$: Do policies trained on BiDemoSyn data transfer zero-shot across different robot embodiments while maintaining high success rates?

\begin{figure*}[h]
	\begin{center}
           \includegraphics[width=0.95\linewidth]{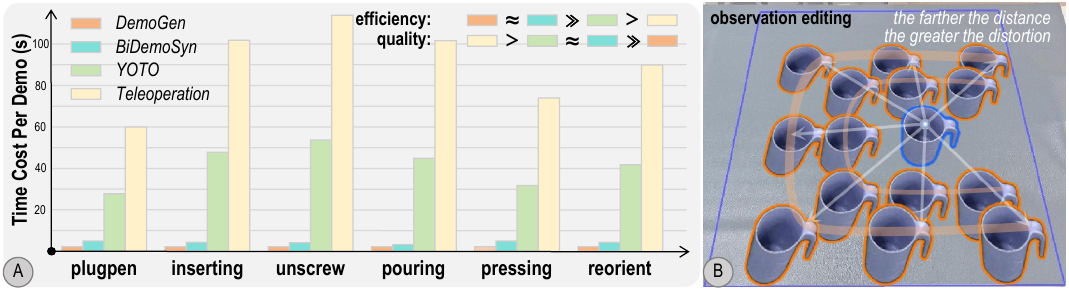}
	\vspace{-5pt}
	\caption{(A) The data collection efficiency comparison of different baselines and our BiDemoSyn, and (B) the generated data quality illustration of DemoGen. Although DemoGen has the highest synthesis efficiency, it cannot avoid visual artifacts caused by perspective transformation, so its data quality is the lowest. Our method can achieve a good speed-quality balance.}
           \label{timecost}
	\vspace{-10pt}
	\end{center}
\end{figure*}

\setlength{\tabcolsep}{0.5pt}
\begin{table*}[!t]\footnotesize  
	\centering
	\caption{Quantitative results of different methods under \textit{in-distribution} (ID) and \textit{out-of-distribution} (OOD) evaluations. The OOD means removing one or a pair of objects from the training set for testing purposes only. The \texttt{\#} means the training size. }
	\vspace{-5pt}
	\begin{tabular}{c|c|cccccc|c|cccccc|c}
	\Xhline{1.2pt}
	~ & ~ & \multicolumn{7}{c|}{\textit{in-distribution} (ID) evaluations} & \multicolumn{7}{c}{\textit{out-of-distribution} (OOD) evaluations} \\
	\cline{3-16}
	\makecell{Policy\\(Input\\Mode)} & Method & \rotatebox[origin=c]{45}{\texttt{plugpen}} & \rotatebox[origin=c]{45}{\texttt{inserting}}
 		& \rotatebox[origin=c]{45}{\texttt{unscrew}} & \rotatebox[origin=c]{45}{\texttt{pouring}}
 		& \rotatebox[origin=c]{45}{\texttt{pressing}} & \rotatebox[origin=c]{45}{\texttt{reorient}}
 		& \makecell{~\\Average\\Success\\Rate} 
		& \rotatebox[origin=c]{45}{\texttt{plugpen}} & \rotatebox[origin=c]{45}{\texttt{inserting}}
 		& \rotatebox[origin=c]{45}{\texttt{unscrew}} & \rotatebox[origin=c]{45}{\texttt{pouring}}
 		& \rotatebox[origin=c]{45}{\texttt{pressing}} & \rotatebox[origin=c]{45}{\texttt{reorient}}
 		& \makecell{~\\Average\\Success\\Rate} \\
	~ & ~ & \texttt{\#3888} & \texttt{\#7776} & \texttt{\#1152} & \texttt{\#2592} & \texttt{\#1296} & \texttt{\#1008} & ~ &
		\texttt{\#2916} & \texttt{\#3888} & \texttt{\#1008} & \texttt{\#972} & \texttt{\#324} & \texttt{\#756} \\
	\Xhline{0.8pt} 
	\multirow{4}{*}{\makecell{Training\\-Free\\(RGB)}} & ReKep & 15/30 & 13/30 & 12/30 & 15/30 & 12/30 & 09/30 & \cellcolor{gray!15}42.2\%  
		& 12/30 & 10/30 & 10/30 & 09/30 & 09/30 & 06/30 & \cellcolor{gray!15}31.1\%   \\  
	~ & ReKep+ & 17/30 & 17/30 & 14/30 & 19/30 & 20/30 & 11/30 & \cellcolor{gray!15}54.4\%   
		& 15/30 & 14/30 & 12/30 & 12/30 & 11/30 & 09/30 & \cellcolor{gray!15}40.6\%   \\  
	~ & ODIL & 18/30 & 19/30 & 13/30 & 18/30 & 18/30 & 12/30 & \cellcolor{gray!15}54.4\%  
		& 12/30 & 14/30 & 08/30 & 11/30 & 13/30 & 08/30 & \cellcolor{gray!15}36.7\%   \\  
	~ & MAGIC & 19/30 & 18/30 & 14/30 & 18/30 & 19/30 & 12/30 & \cellcolor{gray!15}55.6\%  
		& 13/30 & 14/30 & 10/30 & 12/30 & 13/30 & 09/30 & \cellcolor{gray!15}39.4\%   \\  
	\hline 
	\multirow{3}{*}{\makecell{DP\\(RGB)}} & DemoGen & 18/30 & 20/30 & 16/30 & 16/30 & 14/30 & 15/30 & \cellcolor{gray!15}55.0\%  
		& 08/30 & 03/30 & 10/30 & 04/30 & 03/30 & 07/30 & \cellcolor{gray!15}19.4\% \\  
	~ & YOTO & 19/30 & 20/30 & 16/30 & 17/30 & 14/30 & 16/30 & \cellcolor{gray!15}56.7\%  
		& 08/30 & 04/30 & 10/30 & 05/30 & 04/30 & 07/30 & \cellcolor{gray!15}21.1\% \\  
	~ & \textbf{BiDemoSyn} & 22/30 & 24/30 & 22/30 & 21/30 & 17/30 & 19/30 & \textbf{\cellcolor{gray!15}67.8\%} 
		& 13/30 & 10/30 & 18/30 & 15/30 & 08/30 & 12/30 & \textbf{\cellcolor{gray!15}42.2\%} \\  
	\hline 
	\multirow{3}{*}{\makecell{DP3\\(PCD)}} & DemoGen & 21/30 & 24/30 & 19/30 & 20/30 & 18/30 & 17/30 & \cellcolor{gray!15}66.1\%
		& 11/30 & 06/30 & 14/30 & 07/30 & 05/30 & 11/30 & \cellcolor{gray!15}30.0\% \\
	~ & YOTO & 22/30 & 23/30 & 20/30 & 20/30 & 18/30 & 19/30 & \cellcolor{gray!15}67.8\%
		& 12/30 & 06/30 & 15/30 & 08/30 & 06/30 & 14/30 & \cellcolor{gray!15}33.9\% \\
	~ & \textbf{BiDemoSyn} & 26/30 & 28/30 & 25/30 & 24/30 & 21/30 & 22/30 & \textbf{\cellcolor{gray!15}81.1\%} 
		& 16/30 & 14/30 & 21/30 & 18/30 & 11/30 & 18/30 & \textbf{\cellcolor{gray!15}54.4\%} \\
	\hline 
	\multirow{3}{*}{\makecell{EquiBot\\(PCD)}} & DemoGen & 22/30 & 24/30 & 20/30 & 20/30 & 19/30 & 19/30 & \cellcolor{gray!15}68.9\%
		& 13/30 & 10/30 & 16/30 & 11/30 & 10/30 & 13/30 & \cellcolor{gray!15}40.6\% \\
	~ & YOTO & 23/30 & 24/30 & 20/30 & 21/30 & 20/30 & 19/30 & \cellcolor{gray!15}71.1\% 
		& 15/30 & 11/30 & 18/30 & 14/30 & 12/30 & 16/30 & \cellcolor{gray!15}47.8\% \\
	~ & \textbf{BiDemoSyn} & 28/30 & 28/30 & 26/30 & 25/30 & 24/30 & 25/30 & \textbf{\cellcolor{gray!15}86.7\%} 
		& 18/30 & 20/30 & 24/30 & 21/30 & 17/30 & 20/30 & \textbf{\cellcolor{gray!15}66.7\%} \\
	\Xhline{1.2pt}
	\end{tabular}
	\label{tabA}
	\vspace{-15pt}
\end{table*}

\subsection{Experimental Setup and Protocol}

\textbf{Tasks}. We evaluate on six bimanual manipulation tasks requiring contact-rich coordination: \texttt{plugpen}, \texttt{inserting}, \texttt{unscrew}, \texttt{pouring}, \texttt{pressing} and \texttt{reorient}. These tasks cover diverse primitive skills (\textit{e.g.}, grasping, rotating and handover), and involve both rigid and articulated objects. The one-shot demonstration for each task is collected via kinesthetic teaching. For the platform, two fixed-base arms equipped with parallel grippers are used as the main workspace. An auxiliary humanoid dual-arm platform was also used to verify the cross-platform deployment capability of the learned policy. Scene perception is provided by a stereo camera capturing binocular images. More details of tasks, harwares and data collection are in \textit{\textbf{Supplementary Materials}}. 

\textbf{Policies and Baselines}. Compared baselines contain two categories. One category is \textit{purely for data collection} including point cloud editing (DemoGen \cite{xue2025demogen}), real robot auto-rollout (YOTO \cite{zhou2025you}), and human drag teaching (close to Teleoperation). Generally, the quality of collected demonstrations by these baselines is better in turn, but the cost is more time-consuming. After preparing the training data, we adapt three advanced visuomotor policies (DP \cite{chi2023diffusion}, DP3 \cite{ze2024dp3} and EquiBot \cite{yang2024equibot}) to bimanual settings by modifying their modeling spaces to dual-arm actions (refer Sec.~\ref{SecPre}). Observation inputs are RGB-only images or segmented 3D point clouds of task-relevant objects. Final trained policies operate in the open-loop discrete keyposes prediction to align with our synthesized demonstration format. The another category is \textit{directly for bimanual manipulation without retraining} including the zero-shot ReKep \cite{huang2024rekep}, an advanced ReKep+ with oracle-level grasp labels at the beginning, and two one-shot imitation learning methods ODIL \cite{wang2025one} and MAGIC \cite{liu2025one}. 

\textbf{Metrics}. We evaluate (1) \textit{synthesis efficiency}: average time per demo over 50 tests, (2) \textit{data quality}: visual authenticity of newly acquired observations, (3) \textit{success rates}: real robot deployment with 30 trials per task, and (4) \textit{generalization}: performance on unseen object placements and geometries.

\vspace{-5pt}
\subsection{Comparison and Presentation of Results}
We here answer three questions raised earlier, demonstrating that BiDemoSyn can efficiently synthesize high-quality real-world demonstrations, enabling scalable and generalizable visuomotor policy training with minimal human input. Then there are some snapshots of real robot execution effects.

\begin{figure*}[h]
    \centering
    \begin{minipage}{.348\textwidth}
		\centering
		\includegraphics[width=\columnwidth]{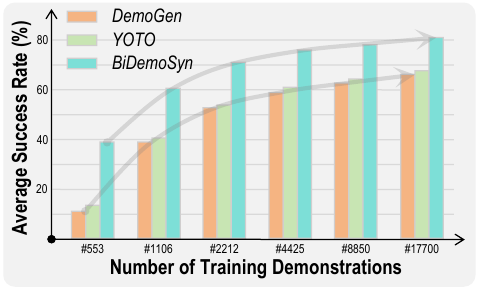}
		\vspace{-18pt}
		\caption{Comparison between training scale and success rate. Less sized data is randomly sampled out of the total dataset at task-level. DP3 is chosen as the visuomotor policy.}
		\label{scalinglaws}
    \end{minipage}
    \hspace{0.02cm}
    \begin{minipage}{.636\textwidth}
		\centering
		\includegraphics[width=\columnwidth]{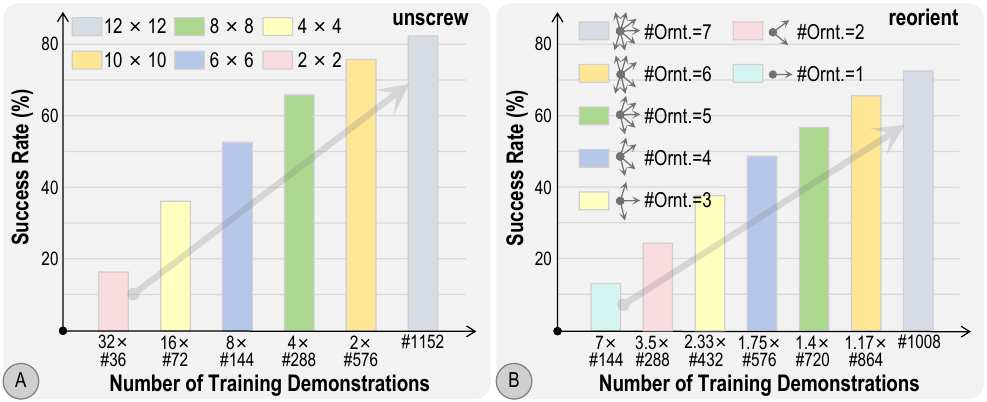}
		\vspace{-18pt}
		\caption{Analysis of spatial generalization for variations in (A) position sampling density and (B) orientation sampling diversity. DP3 is chosen as the policy.}
		\label{generalization}
    \end{minipage}
    \vspace{-8pt}
\end{figure*}

\begin{figure*}[t]
	\begin{center}
           \includegraphics[width=1.0\linewidth]{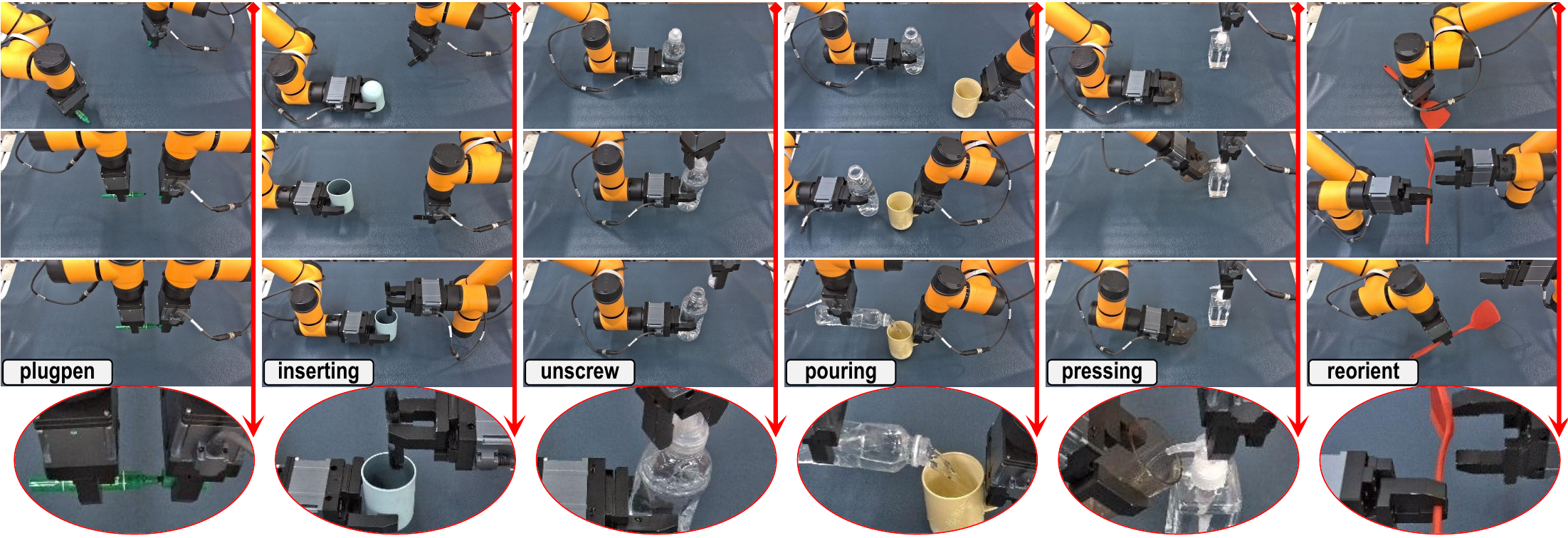}
	\vspace{-18pt}
	\caption{Visualization of all six bimanual tasks performed on real robots. All models are trained and tested under the ID evaluations. EquiBot is chosen as the visuomotor policy. Key highlighted dual-arm coordination movements associated with each task are partially enlarged at the bottom row for quick review.}
           \label{visualization}
	\vspace{-20pt}
	\end{center}
\end{figure*}

($\mathsf{A}1$): \textbf{The efficiency and usability of BiDemoSyn have obvious advantages over baselines}.
Without bells and whistles, our BiDemoSyn demonstrates superior efficiency and usability as illustrated in Fig.~\ref{timecost}. It requires \textit{about 5 seconds per demonstration} to synthesize new trajectories, where primarily involving manual object repositioning and visual verification of initial frame alignment, followed by the trajectory optimization. In contrast, DemoGen generates trajectories in \textit{less then 1 second} via scripted edits but suffers from perspective distortion artifacts, degrading data quality. The specific qualitative effect can be compared with Fig.~\ref{timecost}A and our results in Fig.~\ref{framework}. By relying on manual alignment and auto-replay, YOTO takes \textit{42 seconds per demo} which is equivalent to real robot execution time. While, teleoperation demands \textit{91 seconds per demo} for near-perfect but labor-intensive collection. These results highlight BiDemoSyn’s unique balance of speed (automated optimization) and reliability (artifact-free synthesis), making it the only method scalable to thousands of demonstrations without compromising real-world fidelity.

($\mathsf{A}2$): \textbf{Demonstrations obtained via BiDemoSyn can support scalable imitation learning}.
We applied BiDemoSyn for efficient collection of thousands demonstrations per task, densely covering the workspace with instance-level object diversity, thereby providing abundant training data for imitation learning. For two reproduced baselines, DemoGen \cite{xue2025demogen} matches our data scale but suffers from visual artifacts, while YOTO \cite{zhou2025you} collects only 1/10 the data per task for being limited by its time-consuming replay mechanism. After obtaining adequate data, we trained three advanced visuomotor policies DP \cite{chi2023diffusion}, DP3 \cite{ze2024dp3} and EquiBot \cite{yang2024equibot}. As shown in Tab.~\ref{tabA} left, BiDemoSyn achieves superior \textit{in-distribution} (ID) success rates across all six tasks, outperforming both baselines regardless of the policy architecture. This confirms that BiDemoSyn’s artifact-free synthesis and efficient data scaling reliably support the scaling laws of imitation learning. Besides, it is unsurprising that policies trained with demonstrations consistently outperform three training-free baselines ReKep\cite{huang2024rekep}, ODIL \cite{wang2025one}, and MAGIC \cite{liu2025one}. To further quantify scalability, we evaluated training size-performance relationships on six tasks of all methods. As in Fig.~\ref{scalinglaws}, BiDemoSyn-trained policies exhibit stronger scaling trends than DemoGen (\textit{e.g.}, 60\% vs. 39\% at 1106 demos), aligning with community observations that data quality and quantity jointly drive visuomotor policy performance \cite{liu2025rdt, black2024pi0, pertsch2025fast, lin2025data}.

\setlength{\tabcolsep}{0.5pt}
\begin{table*}[t]\footnotesize  
	\centering
	\caption{Quantitative results for few-shot BiDemoSyn. Although the number of training demonstrations is the same for each setting, the number of seed demonstrations for trajectory synthesis is different. The \texttt{\#} means the training size.}
	\vspace{-5pt}
	\begin{tabular}{c|c|cccccc|c|cccccc|c}
	\Xhline{1.2pt}
	~ & ~ & \multicolumn{7}{c|}{\textit{in-distribution} (ID) evaluations} & \multicolumn{7}{c}{\textit{out-of-distribution} (OOD) evaluations} \\
	\cline{3-16}
	\makecell{Trained\\Policy} & \makecell{Few-Shot\\BiDemoSyn} & \rotatebox[origin=c]{45}{\texttt{plugpen}} & \rotatebox[origin=c]{45}{\texttt{inserting}}
 		& \rotatebox[origin=c]{45}{\texttt{unscrew}} & \rotatebox[origin=c]{45}{\texttt{pouring}}
 		& \rotatebox[origin=c]{45}{\texttt{pressing}} & \rotatebox[origin=c]{45}{\texttt{reorient}}
 		& \makecell{~\\Average\\Success\\Rate} 
		& \rotatebox[origin=c]{45}{\texttt{plugpen}} & \rotatebox[origin=c]{45}{\texttt{inserting}}
 		& \rotatebox[origin=c]{45}{\texttt{unscrew}} & \rotatebox[origin=c]{45}{\texttt{pouring}}
 		& \rotatebox[origin=c]{45}{\texttt{pressing}} & \rotatebox[origin=c]{45}{\texttt{reorient}}
 		& \makecell{~\\Average\\Success\\Rate} \\
	~ & ~ & \texttt{\#3888} & \texttt{\#7776} & \texttt{\#1152} & \texttt{\#2592} & \texttt{\#1296} & \texttt{\#1008} & ~ &
		\texttt{\#2916} & \texttt{\#3888} & \texttt{\#1008} & \texttt{\#972} & \texttt{\#324} & \texttt{\#756} \\
	\Xhline{0.8pt} 
	\multirow{5}{*}{EquiBot} & 1-shot &28/30 & 28/30 & 26/30 & 25/30 & 24/30 & 25/30 & \cellcolor{gray!15}86.7\%  
		& 18/30 & 20/30 & 24/30 & 21/30 & 17/30 & 20/30 & \cellcolor{gray!15}66.7\% \\  
	~ & 5-shot & 28/30 & 28/30 & 26/30 & 26/30 & 25/30 & 25/30 & \cellcolor{gray!15}87.8\%   
		& 20/30 & 21/30 & 24/30 & 23/30 & 18/30 & 21/30 & \cellcolor{gray!15}70.0\% \\  
	~ & 10-shot & 28/30 & 28/30 & 26/30 & 27/30 & 25/30 & 26/30 & \cellcolor{gray!15}88.9\%    
		& 21/30 & 21/30 & 25/30 & 23/30 & 18/30 & 22/30 & \cellcolor{gray!15}71.7\% \\  
	~ & 15-shot & 28/30 & 28/30 & 27/30 & 27/30 & 26/30 & 26/30 & \cellcolor{gray!15}90.0\%    
		& 21/30 & 22/30 & 25/30 & 23/30 & 18/30 & 22/30 & \cellcolor{gray!15}72.2\% \\  
	~ & 20-shot & 28/30 & 28/30 & 27/30 & 27/30 & 26/30 & 26/30 & \cellcolor{gray!15}90.0\%    
		& 21/30 & 22/30 & 25/30 & 23/30 & 18/30 & 22/30 & \cellcolor{gray!15}72.2\% \\  
	\Xhline{1.2pt}
	\end{tabular}
	\label{tabB}
	\vspace{-15pt}
\end{table*}

($\mathsf{A}3$): \textbf{Policies trained on BiDemoSyn data can achieve better generalization to unseen variations}.
To evaluate the generalization to novel instances, we conduct controlled tests across all six tasks. For each task, we exclude demonstrations associated with one or a pair of randomly selected objects (\textit{e.g.}, an arbitrary bottle in \texttt{unscrew} or a bottle-cup pair in \texttt{pouring}) from the training set, resulting in training sizes of 2916/3888/1008/972/324/756 demos for six tasks, respectively. These excluded objects are reintroduced exclusively during real-world testing, with identical data splits applied to DemoGen and YOTO baselines for fair comparison. As shown in Tab.~\ref{tabA} right, policies trained on BiDemoSyn data consistently achieve higher success rates compared to all baselines in \textit{out-of-distribution} (OOD) settings. While all methods exhibit performance drops relative to ID evaluations, BiDemoSyn’s superior OOD robustness highlights its ability to capture task-invariant features through vision-aligned synthesis. To further analyze spatial generalization, we vary the density of positional and orientation coverage in training data: for \texttt{unscrew}, we synthesize trajectories with sparse-to-dense workspace coverage (see Fig.~\ref{generalization}A), and for \texttt{reorient}, we incrementally increase the diversity of object orientations (see Fig.~\ref{generalization}B). To exclude the effect of data size differences, the total number of train-set is always augmented to a standard size (\textit{e.g.}, \texttt{\#1152} and \texttt{\#1008}) via the DemoGen to edit some observations at small distances. Finally, policies trained on BiDemoSyn data with dense spatial coverage (\textit{e.g.}, $10\times10$ workspace coverage in \texttt{unscrew}) achieve 76.7\% success on unseen positions, compared to 53.3\% for sparse coverage ($6\times6$ grid cells). Similarly, orientation diversity in \texttt{reorient} improves generalization from 37.3\% (lower diversity with \#\texttt{Ornt.=3}) to 73.3\% (higher diversity with \#\texttt{Ornt.=7}). These results confirm that BiDemoSyn’s ability to cheaply synthesize positionally and orientationally diverse demos directly translates to enhanced policy generalization, aligning with empirical insights that broad data coverage mitigates spatial distribution shifts.

\textbf{Qualitative Results}:
Fig.~\ref{visualization} shows effectiveness across six tasks. For example, the \texttt{plugpen} policy precisely aligns and inserts the pen into the narrow cap even with $\pm$5mm positional noise. In \texttt{inserting}, dual-arm coordination achieves rough but proximate peg-in-hole precision. The \texttt{unscrew} demonstrates robust rotational synchronization, successfully loosening lids with variable thread tightness. The \texttt{pouring} policy adapts liquid flow control to container geometries, minimizing spills despite tilted orientations. For \texttt{pressing}, predicted actions reliably activate nozzles without over-pressuring, while \texttt{reorient} handles object handovers and flips with smooth dual-arm transitions. These results corroborate our quantitative findings, emphasizing BiDemoSyn’s ability to handle real-world complexity in contact-rich cases.

\begin{figure}[t]
	\begin{center}
           \includegraphics[width=1.0\linewidth]{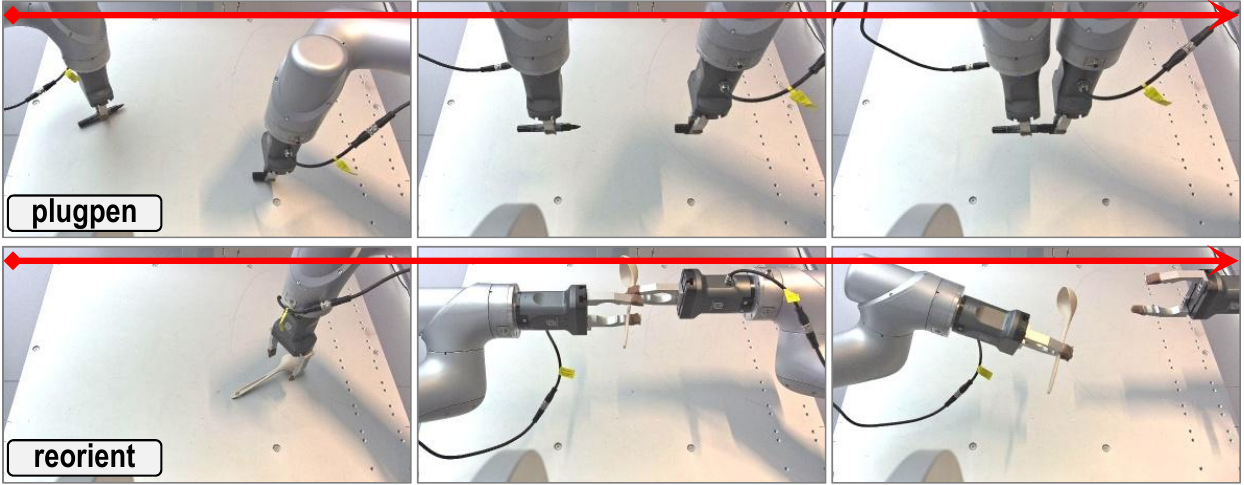}
	\vspace{-15pt}
	\caption{Visualization for the cross-embodiment deployment of BiDemoSyn-trained policies on an unseen humanoid-style dual-arm robot for the \texttt{plugpen} and \texttt{reorient} tasks.}
           \label{rokaeExamples}
	\vspace{-20pt}
	\end{center}
\end{figure}

($\mathsf{A}4$): \textbf{BiDemoSyn can seamlessly integrate with few-shot demonstrations to improve robustness}.
Although BiDemoSyn is formulated in a one-shot paradigm, it can seamlessly incorporate additional real demonstrations (5–20 samples) as complementary seeds without altering the synthesis pipeline. We summarized this ablation results in Tab.~\ref{tabB}. Empirically, we can find that few-shot extensions provide limited gains under ID settings, where one-shot–generated datasets already yield near-saturated performance on seen settings. In contrast, moderate but consistent improvements are observed under OOD conditions, primarily due to expanded coverage of object geometries and intermediate contact states introduced by new demonstrations. These results suggest that BiDemoSyn’s one-shot synthesis already produces high-quality, physically grounded demonstrations, while few-shot seeds mainly contribute by enriching representational diversity rather than correcting systematic failure modes. Overall, this study confirms that BiDemoSyn offers a favorable trade-off: it remains highly effective in the one-shot regime, benefits selectively from few-shot data in OOD scenarios, and supports hybrid one-shot + few-shot usage without increasing complexity.

($\mathsf{A}5$): \textbf{Policies trained on BiDemoSyn data transfer effectively across robot embodiments in a zero-shot manner}.
BiDemoSyn facilitates zero-shot cross-embodiment policy deployment by design, owing to its object-centric observation formulation and the use of end-effector–centric 6-DoF action representation. These choices decouple policy learning from robot-specific kinematics and enable direct transfer to new hardware platforms. To evaluate this capability, we deploy visuomotor policies trained on the primary dual-arm platform to a previously unseen humanoid-style bimanual robot, without additional training or fine-tuning. We consider two representative tasks, \texttt{plugpen} and \texttt{reorient}, using EquiBot policies trained under ID settings. 
As shown in Fig.~\ref{rokaeExamples}, the transferred policies successfully reproduce precise bimanual alignment for insertion and reorientation. Quantitative evaluations indicate that success rates on the new embodiment are comparable to those reported in Tab.~\ref{tabA} left. These results demonstrate that BiDemoSyn-trained policies exhibit strong embodiment-agnostic behavior, highlighting the potential for scalable deployment across heterogeneous robotic platforms.

\section{Conclusion and Limitation}
We introduce BiDemoSyn, a real-world demonstration synthesis framework that enables scalable bimanual manipulation learning from a single kinesthetic example. By decomposing demonstrations into invariant coordination blocks and object-dependent adaptations, and instantiating them via vision-based object anchoring and trajectory optimization, BiDemoSyn generates large-scale, physically feasible demonstrations without simulation or repeated teleoperation. Across diverse bimanual tasks, policies trained on BiDemoSyn data achieve strong generalization to novel objects, long-horizon compositions, and unseen robot embodiments. We further show that the framework naturally extends to few-shot synthesis, improving robustness with minimal additional supervision. This work establishes a new paradigm for scalable imitation learning, bridging the divide between data quantity and physical fidelity in bimanual robotic manipulation.

\textbf{Limitations}: While BiDemoSyn excels in static, geometrically varied scenes, it faces challenges in fully dynamic environments and extreme shape variations (\textit{e.g.}, highly deformable or articulated objects). The current hierarchical optimization assumes quasi-static motions, limiting its applicability to highly dynamic tasks like catching. Additionally, synthesizing trajectories for multi-stage tasks with interdependent contacts (\textit{e.g.}, lacing ropes) requires further extension. Future work will integrate dynamic perception and adaptive contact modeling to address these constraints.


\section*{Acknowledgments}
This work was supported by the Key-Area Research and Development Program of Guangdong Province, China under Grant 2024B0101040004, the Shenzhen Science and Technology Program under Grant KJZD20240903104008012, and the Shenzhen Science and Technology Program under Grant ZDCY20250901113000001.

\bibliographystyle{plainnat}
\bibliography{refs}

\newpage

\section*{Appendix}


This supplementary provides detailed explanations and extensions to the main paper. Sec.~\ref{appdA} outlines the design rationale of six bimanual manipulation tasks, including platform specifications, task-specific objectives, and one-shot demonstration acquisition protocols. Sec.~\ref{appdB} details the data collection standards, UI-assisted collection tools, and post-processing workflows. These constitute the trajectory synthesis pipelines for generating real-world demonstrations with BiDemoSyn. We have also added more details about BiDemoSyn to help to understand it. Sec.~\ref{appdC} elaborates on policy training, covering implementation specifics for baselines (DemoGen \cite{xue2025demogen}, YOTO \cite{zhou2025you}) and diffusion policies (DP3 \cite{ze2024dp3}, EquiBot \cite{yang2024equibot}), alongside real-robot deployment. Sec.~\ref{appdD} provides additional visualizations of real-robot executions and failure case analyses, offering insights into current limitations. Sec.~\ref{appdE} concludes with reflections and future directions, aiming to extend BiDemoSyn to challenge more complex tasks and enhance cross-domain generalization.

\section{Design of Bimanual Manipulation Tasks}\label{appdA}

\subsection{Hardware and Platform}
\begin{figure}[h]
	\begin{center}
           \includegraphics[width=\linewidth]{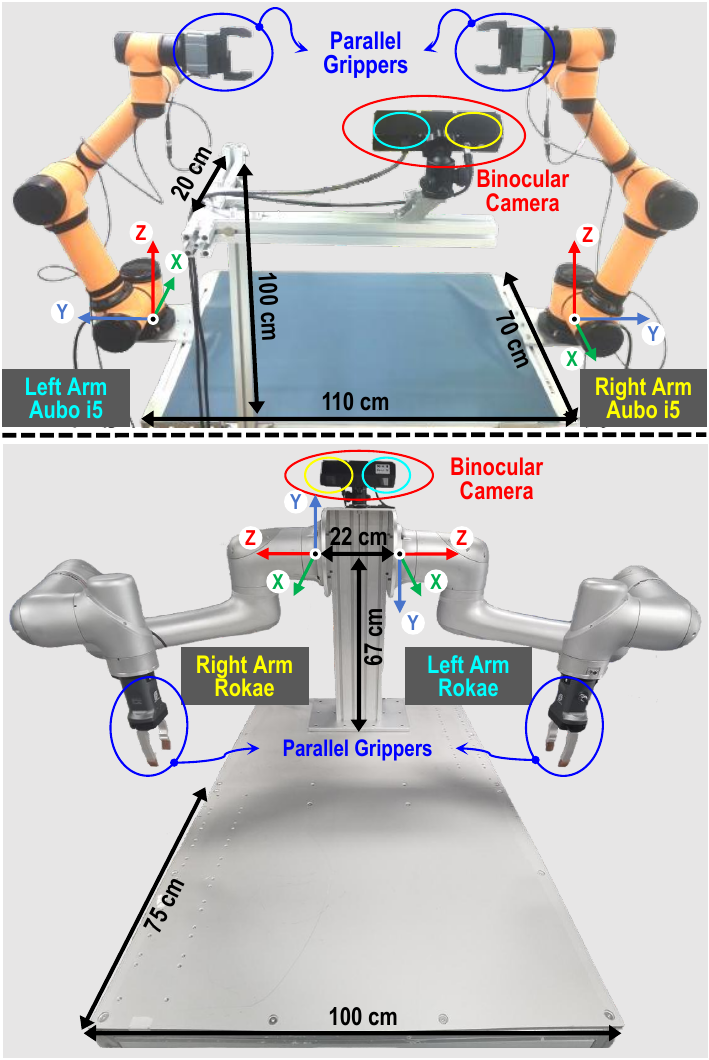}
	\vspace{-15pt}
	\caption{\textit{(Top) Primary}. The fixed-base dual-arm platform having a table with two robot arms, two grippers and the binocular camera. \textit{(Bottom) Auxiliary}. The humanoid dual-arm platform for validating cross-embodiment capabilities.}
           \label{platform}
	\vspace{-20pt}
	\end{center}
\end{figure}

Our main experimental platform comprises a rectangular workspace (110cm$\times$70cm) with two 6-DoF AUBO-i5 collaborative arms\footnote{https://www.aubo-cobot.com/public/i5product3} (880mm reach) mounted on opposite short edges of the table (refer Fig.~\ref{platform} top). This opposing-arm configuration maximizes shared workspace while minimizing self-collision risks, albeit differing from anthropomorphic designs. Each arm is equipped with a DH-Robotics parallel gripper\footnote{https://en.dh-robotics.com/product/pg} (80mm max opening, 50mm effective length), controlled in binary states (open/closed). Tool length compensation accounts for 160mm absolute length of the gripper. The scene perception is provided by a binocular stereo Kingfisher R-6000 camera (960$\times$540 RGB resolution), mounted 100cm above the table long edge to capture a third-person view of the workspace. The calibrated stereo setup reconstructs high-fidelity 3D point clouds \cite{xu2023iterative}, eliminating the need for wrist-mounted cameras while ensuring full task visibility. To further demonstrate the cross-embodiment transferability of trained policies based on BiDemoSyn, as shown in Fig.~\ref{platform} bottom, we have prepared another new dual-arm robotic platform configured in a popular humanoid style. This new platform consists of two Rokae xMate CR7\footnote{https://www.rokae.com/en/product/show/545/xMateCR.html} 6-DoF collaborative arms (reach: 988 mm), each equipped with a parallel gripper (Jodell Robotics RG75-300\footnote{https://www.jodell-robotics.com/product-detail?id=5}, max opening: 75 mm). A binocular camera Kingfisher R-6000 is mounted centrally at the head position. We will present how to utilize this dual-arm platform in Sec.~\ref{appdDrokae} to deploy the learned models and perform real robot testing.

\subsection{Bimanual Task Formulation}
We design six bimanual tasks (\texttt{plugpen}, \texttt{inserting}, \texttt{unscrew}, \texttt{pouring}, \texttt{pressing} and \texttt{reorient}) to comprehensively evaluate dual-arm coordination and generalization. Each task involves at least two category-level object instances (Fig.~\ref{assets}) and requires diverse primitive skills, such as single-arm actions (grasping, placing, precision rotation) and dual-arm coordination (fixed-twisting, plug-in, handover). All tasks are inherently \textbf{long-horizon}, starting from goal-conditioned initial grasping (objects fully separated from robots), contrasting prior works already grasping/holding the manipulated object and focusing on short-horizon atomic skills such as untwisting \cite{lin2024twisting, bahety2024screwmimic} or handover \cite{huang2023dynamic}. Fig.~\ref{teaser} shows some examples. Below we detail each task:

\begin{figure}[t]
	\begin{center}
           \includegraphics[width=\linewidth]{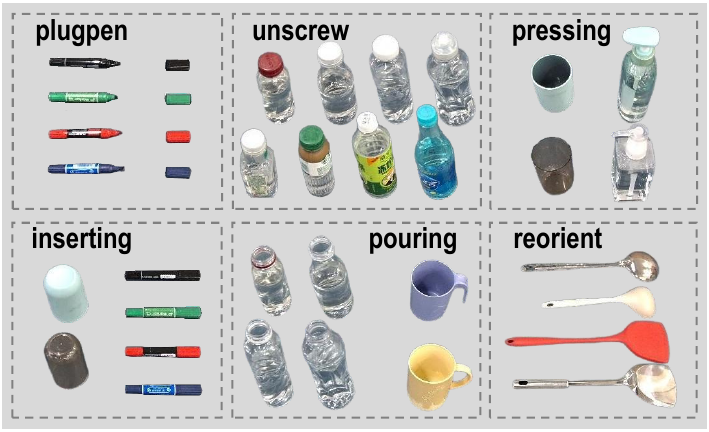}
	\vspace{-15pt}
	\caption{Object assets involved in six bimanual manipulation tasks. All objects have been scaled down proportionally.}
           \label{assets}
	\vspace{-20pt}
	\end{center}
\end{figure}


\begin{itemize}
\item \texttt{plugpen}: \textit{plug the pen with the pen cap}. \textbf{States}: A marker pen body and its cap are placed separately on the table. The pen body lies horizontally with its tip roughly pointing to the right arm, while the cap is also positioned horizontally with its socket facing the left arm. \textbf{Steps}: Left arm grasps pen body; right arm grasps cap. Arms lift and spatially align the pen tip with the cap's socket before insertion. \textbf{Challenges}: Directional grasping alignment, sub-millimeter positional tolerance for insertion.
\vspace{-2pt}
\item \texttt{inserting}: \textit{insert the marker pen into the cup}. \textbf{States}: An empty handle-free cup is placed upside-down on the table, and a closed marker pen lies horizontally nearby. \textbf{Steps}: Left arm flips the cup (about 180$^{\circ}$ rotation); right arm reorients the marker vertically. Arms coordinate to align and insert the marker into the upright cup. \textbf{Challenges}: Dual-arm rotational synchronization, tight insertion tolerance ($\pm$2mm).
\vspace{-2pt}
\item \texttt{unscrew}: \textit{open the bottle by twisting cap counterclockwise}. \textbf{States}: A translucent plastic bottle (filled with colorless water) stands upright on the table, with its cap tightly screwed on. \textbf{Steps}: Left arm stabilizes the bottle; right arm grasps the cap and rotates it vertically (with multiple degree-fixed rotations). \textbf{Challenges}: Force-agnostic cap grasping (no torque sensing), rotational precision ($\pm$5$^{\circ}$ per turn).
\vspace{-2pt}
\item \texttt{pouring}: \textit{pour water from the bottle into the mug cup}. \textbf{States}: An uncapped water bottle (about $3/4$ full) and an empty handled mug are placed on opposite sides of the table. \textbf{Steps}: Left arm tilts the bottle (about 90$^{\circ}$); right arm positions the mug to catch water. \textbf{Challenges}: Fluid dynamics approximation, nozzle-cup alignment under gripper deflection.
\vspace{-2pt}
\item \texttt{pressing}: \textit{press bottle and catch water using the cup}. \textbf{States}: A shampoo bottle (with a pressable nozzle) containing water and an upright empty handle-free cup are placed apart. \textbf{Steps}: Right arm vertically presses the nozzle; left arm tilts the cup to catch water. \textbf{Challenges}: Nozzle positioning accuracy ($\pm$5mm), open-loop fine-grained force control.
\vspace{-2pt}
\item \texttt{reorient}: \textit{flip the spoon so that the bottom is facing up}. \textbf{States}: A long spoon or shovel (metal or plastic) lies concave-down on the table, with randomized orientation and position. \textbf{Steps}: Right arm grasps the spoon center, reorients it mid-air, and hands it to the left arm, which then flips and places it. \textbf{Challenges}: Unstable grasp during handover, rotational precision for flipping ($\pm$10$^{\circ}$).
\end{itemize}

\begin{figure*}[h]
	\begin{center}
           \includegraphics[width=\linewidth]{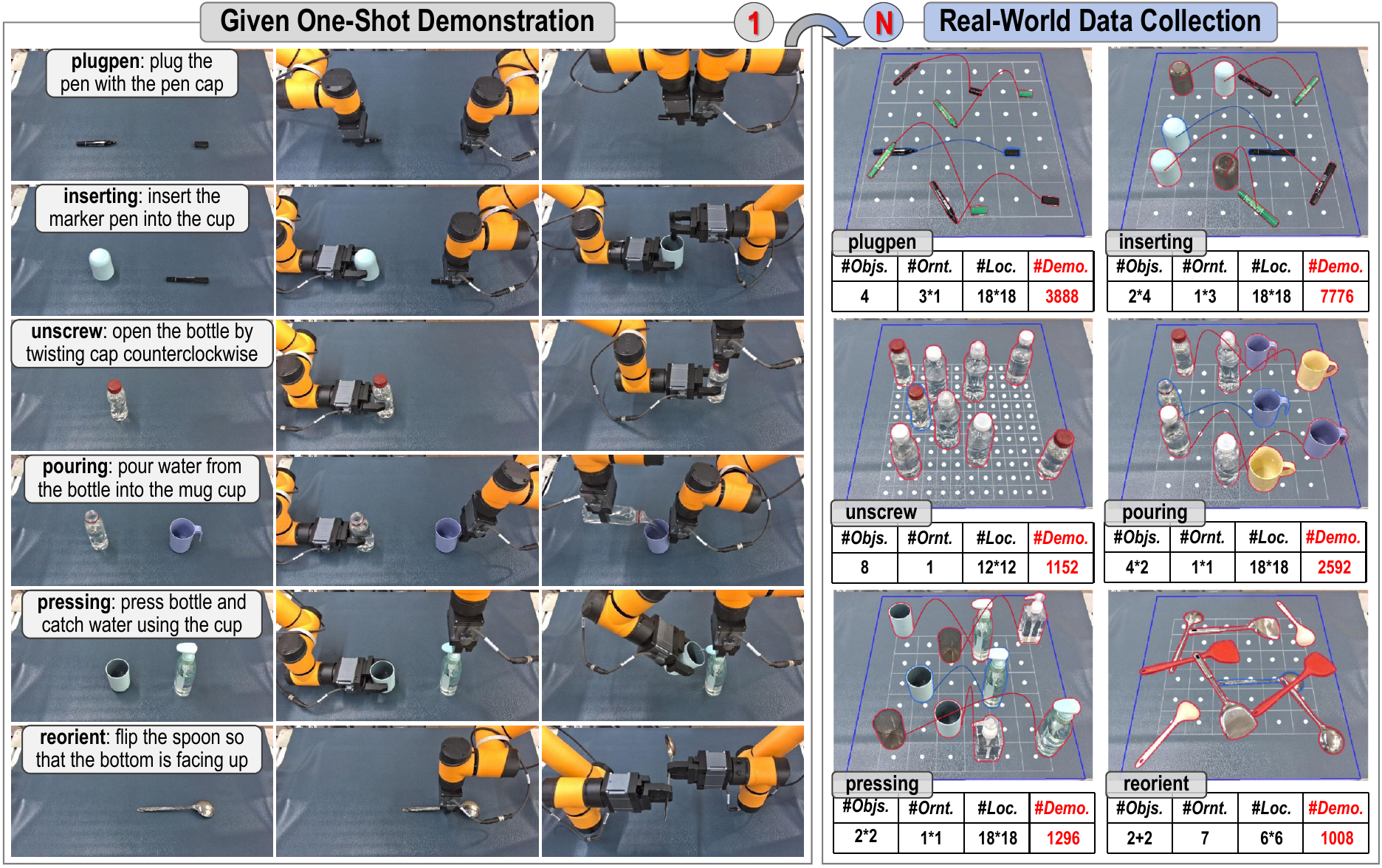}
	\vspace{-15pt}
	\caption{\textbf{From One to Many \textcircled{\scriptsize 1}$\rightarrow$\textcircled{\scriptsize N}}. \textit{Left}: Six representative bimanual manipulation tasks with their one-shot demonstrations and task-specific descriptors. \textit{Right}: Real-world data collection diagrams, showing object instances with varied geometries and spatial arrangements used to synthesize diverse demonstrations (\textit{e.g.}, thousands physically consistent trajectories per task). }
           \label{teaser}
	\vspace{-20pt}
	\end{center}
\end{figure*}

These tasks collectively stress \textit{spatial reasoning}, \textit{contact-aware coordination}, and \textit{generalization} to \textit{instance-level variations}, which are cornerstones of real-world bimanual manipulation. Besdies, it is important to note that, we have \textbf{shared hardware constraints}: (1) No force/torque sensing limits contact-rich adjustments (\textit{e.g.}, pen plug-in force, cap twisting force, or pressing force). (2) Binary gripper states (open/closed) restrict fine-grained manipulation (\textit{e.g.}, plastic bottles are always deformed by excessive clamping). (3) Third-person view occlusions occasionally hinder precise alignment (\textit{e.g.}, cannot discern the groove side of the pen cap).

\subsection{Obtaining and Decomposing the One-Shot Demonstration}
As stated in the main content, we collect one-shot demonstrations via kinesthetic teaching: an operator manually guides both arms through task-critical waypoints, recording the 6DoF end-effector poses (relative to each robot base frame) and gripper binary states at each pause. Objects are placed in fixed initial configurations (allowing small positional tolerance) to ensure consistency. The recorded waypoints are then executed autonomously by the robot control API, which solves inverse kinematics between consecutive given poses and synchronizes gripper actions (\textit{e.g.}, closing after reaching a pre-grasp pose). During auto-execution, stereo camera observations (10Hz) and dual-arm joint/end-effector states are logged. For the real rollout effect of the one-shot demonstration related to each task, please refer to our \textbf{Supplementary Videos}. Finally, demonstrations are deconstructed into task-aware blocks to support subsequent trajectory synthesis.

To enable reliable downstream synthesis, the one-shot kinesthetic demonstration must be collected in a way that naturally exposes task-relevant structure. During kinesthetic teaching, the demonstrator physically guides the two manipulators through the task while pausing at intuitive semantic boundaries, such as after a stable grasp, after completing an alignment adjustment, or before switching from transportation to interaction. These pauses, which arise organically from human motion patterns, produce clear temporal discontinuities that align well with the block-based representation introduced in the main paper. Importantly, this process \textit{does not require any manual annotation or labeling}. The demonstration is captured in a continuous stream exactly as in standard kinesthetic teleoperation, and the boundaries between coarse blocks emerge directly from the dynamics of human-guided motion. This makes the procedure no more labor-intensive than collecting a single ordinary demonstration.

We explored alternative strategies for fully automatic segmentation, such as keyframe/keypose detection and curvature-based change-point identification \cite{james2022coarse, shridhar2023perceiver, ma2024hierarchical}. However, reliably identifying semantically meaningful waypoints from a single demonstration proved brittle, especially when waypoints must correspond to subtle but crucial phases of contact establishment, re-grasping, or dual-arm synchronization. Hardware-assisted approaches (e.g., instrumented gloves as in DexCap \cite{wang2024dexcap} or end-effector teaching interfaces as in UMI \cite{chi2024universal} and DexUMI \cite{xu2025dexumi}) can offer additional cues for segmentation but introduce alignment overhead, cross-morphology calibration challenges, and potential loss of semantic correspondence between human motion and robot embodiment. Given these trade-offs, the kinesthetic-teaching-driven segmentation remains a pragmatic and robust choice: it imposes no additional annotation burden, preserves semantic alignment across blocks, and provides boundaries well suited for constructing invariant and variant components for real-world synthesis.

\section{Details of Data Collection and Process}\label{appdB}

\begin{figure}[t]
	\begin{center}
           \includegraphics[width=\linewidth]{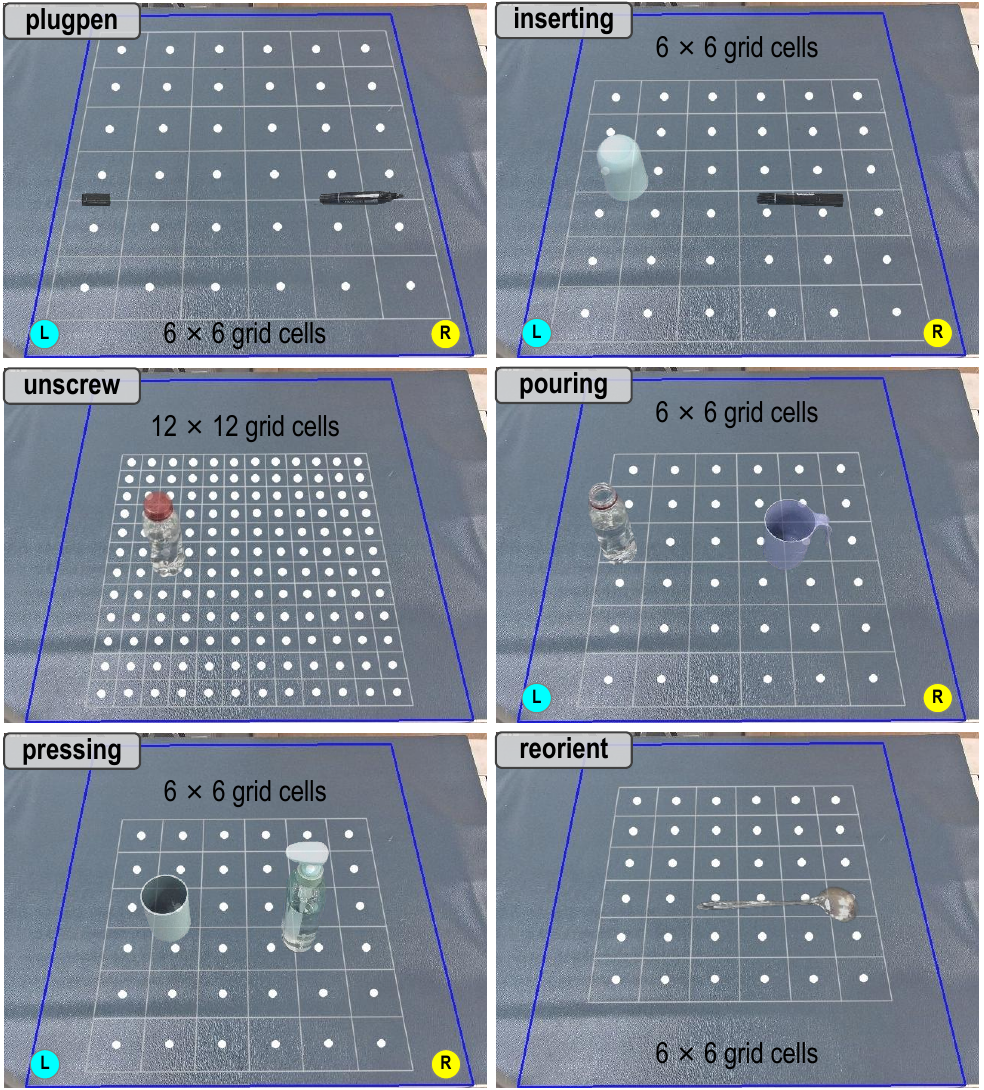}
	\vspace{-15pt}
	\caption{The grid cell division for each task. For tasks involving two manipulated objects, the total number of grid cells will be divided equally between the left and right side.}
           \label{gridcells}
	\vspace{-20pt}
	\end{center}
\end{figure}

\subsection{Task-Oriented Collection Standards}\label{appdB1}
To systematically study factors influencing policies, we establish task-specific data collection protocols that ensure spatial uniformity and instance balance. Each task defines a common workspace (blue quadrilateral in Fig.~\ref{gridcells}, mapped to a 615mm$\times$675mm rectangular area on the table) where objects are placed. For single-object tasks (\textit{e.g.},  \texttt{unscrew}), objects are positioned across grid cells (white dots/boxes in Fig.~\ref{gridcells}) covering the entire workspace. For dual-object tasks, the workspace is split into left-right regions (\textit{e.g.}, bottles occupy the left half, mugs the right half in \texttt{pouring}), with objects distributed uniformly within their zones. Object orientations are randomized for \texttt{plugpen}, \texttt{inserting}, and \texttt{reorient} but fixed for  \texttt{unscrew}, \texttt{pouring}, and \texttt{pressing} (see Fig. 1 in the main paper for orientation counts). A grid cell is marked as collected if its centroid hosts a demonstration, allowing local variations with positional tolerance. This protocol balances coverage and practicality, enabling controlled studies on spatial and instance-level generalization.

\begin{figure*}[t]
	\begin{center}
           \includegraphics[width=\linewidth]{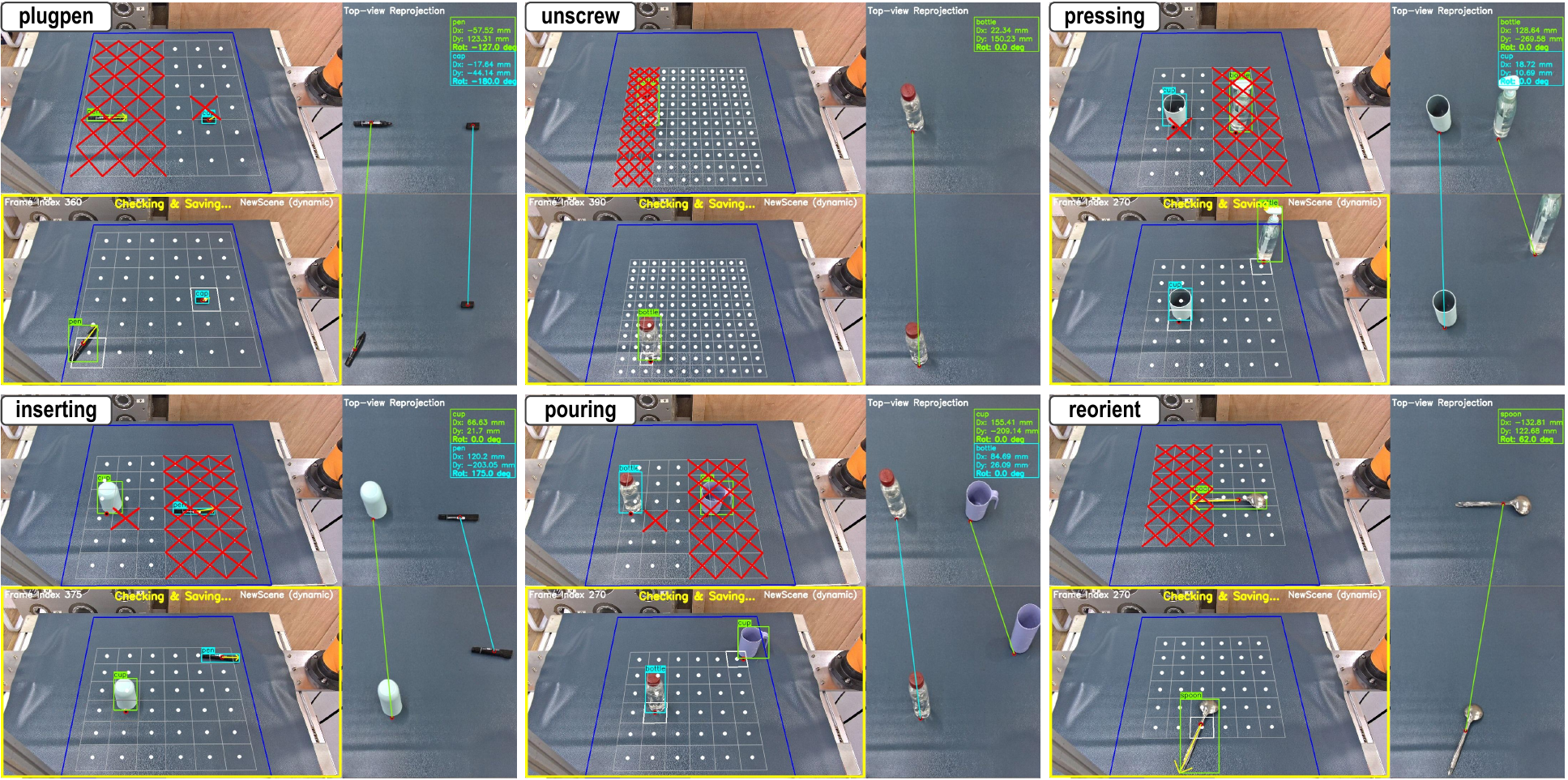}
	\vspace{-15pt}
	\caption{UI interfaces. When collecting data, six tasks need to display different grid cells in real time, as well as perceive and track task-related objects. Note that these points, crosses and lines are drawn digitally, which are not marks in the real world. }
           \label{interface}
	\vspace{-20pt}
	\end{center}
\end{figure*}

\subsection{Collection UI Interface Development}
To streamline data collection, we develop a real-time UI interface (Fig.~\ref{interface}) that visualizes task-specific grid cells, dynamically tracks object positions/poses, and validates alignment with the one-shot demonstration. The interface overlays grid layouts onto live camera feeds, highlights detected objects (using YOLOE \cite{wang2025yoloe} for fast detection and segmentation), and marks completed cells using red crosses. The position of objects is represented by red dots (some objects also use yellow arrows to indicate their orientation). In addition, for cups and bottles that do not need to pay attention to changes in orientation, we did not use the mask centroid as their representive point during collection, but instead used the lowest endpoint in contact with the desktop. A sidebar displays positional/orientational deviations from the reference demonstration, enabling quick verification. If object perception fails, the system falls back to Florence2 \cite{xiao2024florence} and SAM2 \cite{ravi2024sam} for robust detection and segmentation. Operators trigger a \textit{capture-align-validate-save} cycle every 2 seconds. And successful saves prompt visual feedback (UI flash of corresponding grid cells), signaling readiness for the next sample. This setup eliminates manual grid marking and minimizes human idle time, achieving a high speedup while ensuring spatial consistency. Please refer our \textbf{Supplemental Videos} that demonstrate the vivid workflow, illustrating how the UI balances efficiency with precision.

\subsection{Post-Processing of Collected Data}
This part includes the last two stages of BiDemoSyn: initial frame alignment and trajectory optimization. First, pre-recorded positional/orientational deviations (validated via the UI interface during collection) align novel observations with the one-shot demonstration reference frame using the rigid transformation. Next, trajectory optimization ensures kinematic feasibility. We prune singular configurations and validate collision-free paths using the robot IK solvers and motion planners. Over 96\% of trajectories pass validation, with failures (\textit{e.g.}, unreachable corner grid cells) either re-collected or discarded. Finally, validated trajectories are stored in $\mathcal{D}_{syn}$ to complete the synthesis pipeline.

\subsection{Hyperparameters and Predefined Symbols}
In the method section, several thresholds and parameters are introduced for clarity and reproducibility. Here we further explain their meaning and selection strategy. For instance, threshold symbols $\delta$, $\zeta$ and $\gamma$ in Eqn.~4, Eqn.~5 and Eqn.~10 are predefined only for one-shot deconstruction. The symbol $\delta$ is a minimal motion saliency threshold, used to identify boundaries between coarse blocks. The symbol $\zeta$ is a static tolerance threshold to segment motion boundaries under quasi-static assumptions. The symbol $\gamma$ is a refinement threshold used to create fine-grained AEPs after bounding block duration ($T_\textrm{\scriptsize max}$). And the scaling factor $\lambda$ in Eqn.~10 controls the adaptation from the canonical one-shot demonstration to new object instances with varying sizes. Theoretically, $\lambda \in \mathbb{R}^{+}$, yet empirical tuning shows that values within $[0.8, 1.0]$ are sufficient for stable grasp transfer without destabilizing contact. Larger or smaller values tend to shift grasp points away from functional regions (e.g., bottle mid-section in \texttt{unscrew}), while $\lambda=1.0$ already suffices for most tasks. Similarly, thresholds for collision-checking or pose-alignment rejection (e.g., rejecting infeasible IK solutions or near-singular samples) were determined through small-scale grid search and validated in pilot runs. Importantly, such hyperparameters are not per-trial tunable knobs, but rather fixed system constants, ensuring consistent data synthesis across tasks.

\subsection{Evaluation of Intermediate Processes}
We provide further evaluation to justify intermediate design choices from three aspects: Segmentation Quality, Pose Estimation Robustness and Impact on Data Reliability. (1) Firstly, object masks were obtained from Florence2 \cite{xiao2024florence} and SAM2 \cite{ravi2024sam}, two state-of-the-art vision foundation models (VFMs). On held-out captures without foreground occlusion, the segmentation accuracy exceeded 97\%, ensuring reliable geometric inputs for subsequent processing. (2) Secondly, the axis alignment for objects with a distinct orientation was based on classical image-moment theory: first-order moments define centroids, second-order moments define orientation. This method, given reliable masks, yields highly stable results. Quantitative inspection of $\sim$200 sampled trials revealed $<$2\% pose estimation errors, and only under severe occlusion. (3) Thirdly, in Sec.~\ref{appdDfailure}, we showed that perception and orientation errors together account for $\sim$47\% of failures, highlighting their importance. Nevertheless, these intermediate modules are already close to optimal under current open-loop constraints, which explains the high success rate ($\sim$98\%) of the data collection system.

\subsection{Positioning with Respect to Prior Works}
To avoid potential confusion with prior related works such as ReKep \cite{huang2024rekep}, ODIL \cite{wang2025one} and MAGIC \cite{liu2025one}, we emphasize both the similarities and the fundamental differences. Like these methods, BiDemoSyn exploits one-shot demonstrations and compositional reasoning; however, \textit{its core objective is not direct policy execution}, but rather \textit{rapid and scalable synthesis of real-world training demonstrations}. This distinction is crucial: while prior methods target zero-/few-shot policy generalization (often trading reliability for flexibility), BiDemoSyn deliberately focuses on maximizing data quality and efficiency, enabling downstream training of robust visuomotor policies (e.g., DP3 \cite{ze2024dp3} and EquiBot \cite{yang2024equibot}). Our evaluation against DemoGen \cite{xue2025demogen} and YOTO \cite{zhou2025you}, which are two strong baselines addressing the same bottleneck, demonstrates that BiDemoSyn achieves a superior balance between efficiency and data reliability, which is the practical need for scaling imitation learning.

A concurrent work MoMaGen \cite{li2025momagen} tackles scalable bimanual mobile manipulation by generating diverse trajectories in \textit{simulation}, focusing on base reachability and camera visibility through constrained optimization. Although sharing a high-level goal of reducing human effort, MoMaGen and BiDemoSyn operate under fundamentally different assumptions: MoMaGen relies on noise-free simulation assets and virtual scene perturbations, whereas BiDemoSyn performs \textit{real-world demonstration synthesis}, directly addressing physical-domain challenges such as noisy perception, contact feasibility, and one-shot decomposition under real sensor observations. As a result, MoMaGen expands diversity through simulated scene variation, while BiDemoSyn enables \textit{physically grounded generalization} across spatial and category-level object variations without sim-to-real transfer. Thus, BiDemoSyn should be viewed not as a direct competitor to simulation-based frameworks like MoMaGen, but as a \textit{complementary real-world solution} specifically designed for scalable, high-fidelity data generation where simulation pipelines cannot fully capture real-world physical constraints.

\subsection{Evaluation of Demonstration Quality}
Evaluating the quality of synthesized demonstrations is essential because the downstream visuomotor policies depend on both the fidelity of observations and the physical feasibility of actions. We assess demonstration quality along two orthogonal dimensions that align with the structure of visuomotor imitation learning:

\textbf{Visual Fidelity}. This dimension evaluates whether the observations included in a synthesized demonstration faithfully reflect the real physical scene without introducing artifacts. Teleoperation and BiDemoSyn preserve raw RGB-D observations, ensuring perfect consistency with the physical environment. YOTO \cite{zhou2025you} and DemoGen \cite{xue2025demogen} modify the seed point cloud to reposition objects, which introduces geometric distortions and depth inconsistencies, particularly as the displacement grows (see Fig.~4B in the main paper). These artifacts degrade the policy's perception module by coupling unrealistic geometry with real sensor noise.

\textbf{Trajectory Feasibility \& Physical Grounding}. A demonstration is considered high-quality if the associated action sequence: (1) yields a collision-free, dynamically stable trajectory, (2) respects the contact geometry implied by the task semantics, and (3) can be executed on the real robot without additional correction or replanning. Teleoperation and YOTO \cite{zhou2025you} provide ground-truth trajectories. BiDemoSyn produces partially reused (invariant blocks) and partially adapted (variant blocks) trajectories aligned with real-world object pose and geometry. DemoGen \cite{xue2025demogen} synthesizes trajectories directly from edited scene representations, which can deviate from physically stable contact geometries.

Finally, Fig.~4A in the main paper reflects the interplay of these two components. Demonstrations with high-quality observations and physically feasible actions rank highest. This ordering also predicts downstream policy performance: policies trained with cleaner, physically grounded demonstrations demonstrate higher success rates and better robustness under OOD evaluation. These expanded analyses clarify the qualitative and quantitative factors driving the reported comparisons and support the role of demonstration quality as a key determinant of real-world imitation learning performance.

\subsection{Diversity of Synthetic Demonstrations}
A natural question is whether additional diversity (such as reordering or randomizing block sequences) would improve the effectiveness of the synthesized dataset. In BiDemoSyn, we intentionally do not introduce such sequence-level variations. Our empirical findings align with recent analyses in large-scale teleoperation datasets focusing on the diversity exploration and ablation for scalable robotic manipulation \cite{shi2025diversity}, which show that low-level behavioral variability (e.g., minor differences in grasp location, end-effector speed profiles, or redundant micro-motions) does not necessarily benefit policy learning and may even hinder convergence. Instead, we adopt the principle that \textit{task success is primarily determined by achieving correct interactions, not by how those interactions are executed}.

Accordingly, in BiDemoSyn's design, we maintain the original causal structure of the task by preserving the block ordering extracted from the one-shot demonstration, while introducing diversity along the \textit{dimensions that matter for generalization}, namely: (1) spatial variation of object pose (translation + rotation), and (2) category-level instance variation (shape and size differences within the same object class). These forms of variation are directly tied to visuomotor generalization and produce clear benefits in downstream policy performance. In contrast, randomizing the internal block structure introduces behavioral diversity that is orthogonal to task semantics and does not improve success rates in our preliminary tests. We therefore avoid unnecessary variation and instead focus on \textit{environment-induced diversity}, which is both meaningful and consistent with the underlying task geometry.

\begin{figure}[t]
	\begin{center}
           \includegraphics[width=\linewidth]{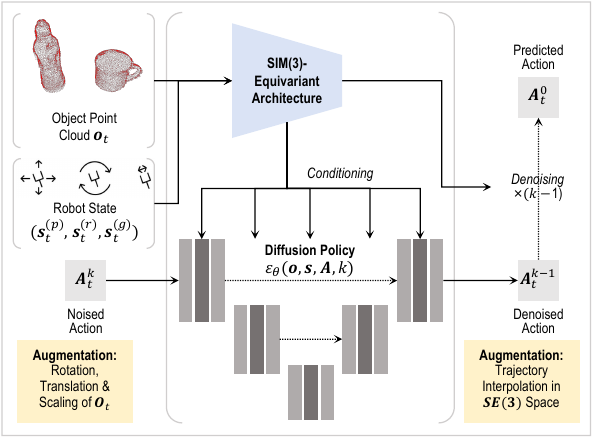}
	\vspace{-15pt}
	\caption{The adapted visuomotor network architecture for imitation learning of bimanual tasks. The input visual observation is simplified to a point cloud of manipulated objects. Note that the vision encoder is slightly different in DP3 \cite{ze2024dp3} and EquiBot \cite{yang2024equibot}. The illustrated one is for EquiBot. }
           \label{networkDP}
	\vspace{-20pt}
	\end{center}
\end{figure}

\begin{figure*}[t]
	\begin{center}
           \includegraphics[width=\linewidth]{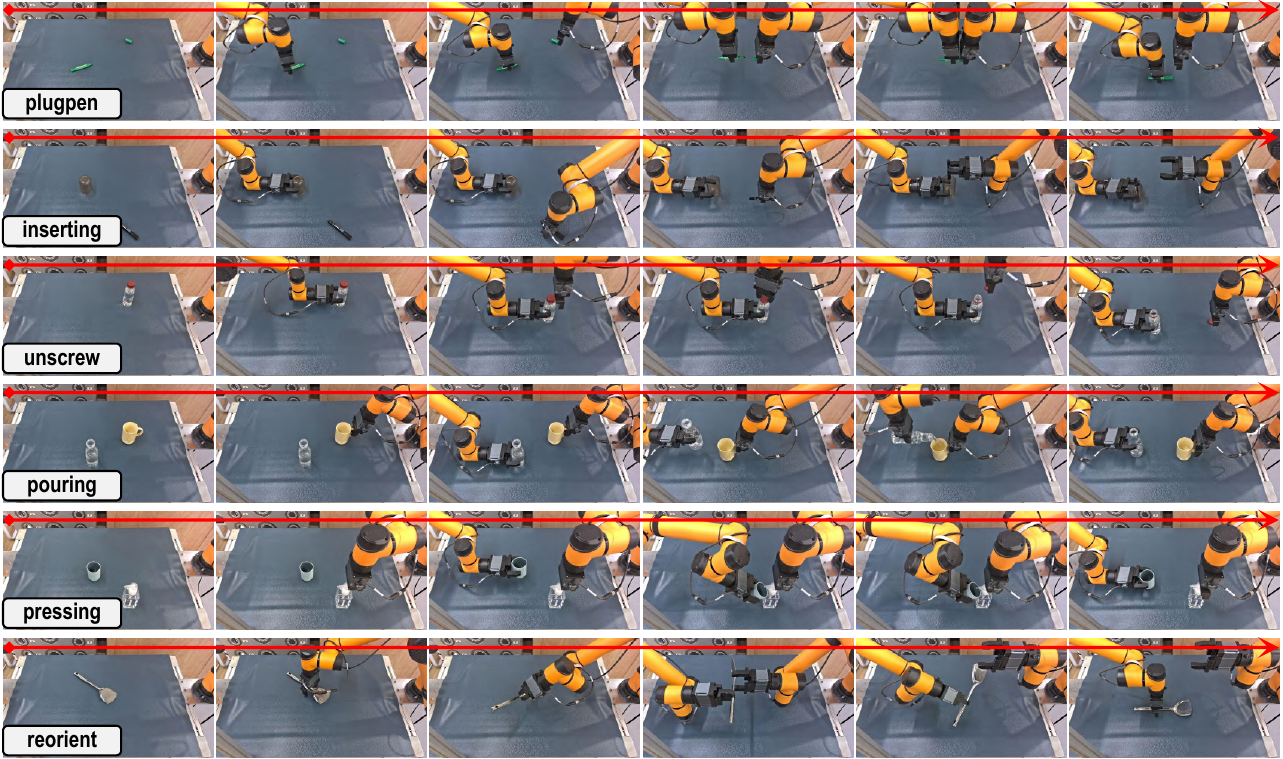}
	\vspace{-15pt}
	\caption{Qualitative real robot rollout samples for defined six bimanual tasks. Best to view after zooming in. }
           \label{moreVis}
	\vspace{-15pt}
	\end{center}
\end{figure*}

\begin{figure*}[h]
	\begin{center}
           \includegraphics[width=1.0\linewidth]{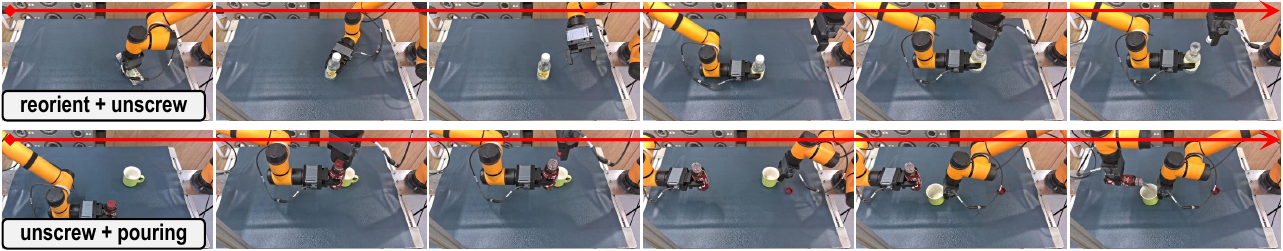}
	\vspace{-15pt}
	\caption{Qualitative real robot rollout examples for two newly added long-horizon tasks (\texttt{pouring} and \texttt{reorient+unscrew}).}
           \label{newTasks}
	\vspace{-20pt}
	\end{center}
\end{figure*}

\begin{figure*}[t]
	\begin{center}
           \includegraphics[width=1.0\linewidth]{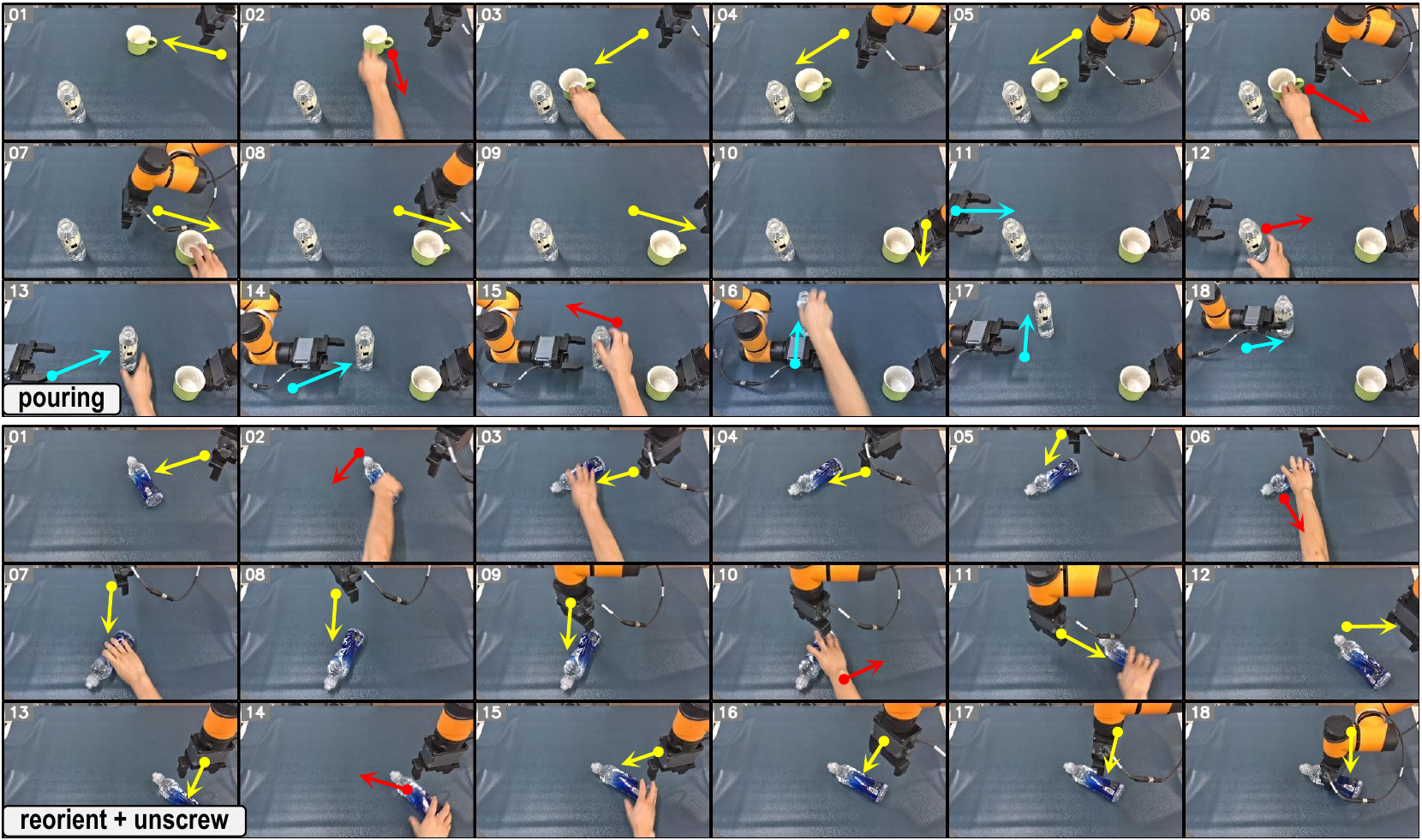}
	\vspace{-15pt}
	\caption{Qualitative examples of applying dynamic interferences during the pre-grasping execution. From top to bottom, they are segments of the dynamic closed-loop grasping phase of tasks \texttt{pouring} and \texttt{reorient+unscrew}, where each object is manually disturbed from one to three times. The red arrow indicates the direction of the manually moved object (interfering). The cyan arrow and yellow arrow indicate the movement direction of the left and right robotic arms (chasing), respectively.}
           \label{dynaVis}
	\vspace{-20pt}
	\end{center}
\end{figure*}

\section{Training and Deployment of Imitation Policies}\label{appdC}

\subsection{Reproduction of Baselines}
For DemoGen \cite{xue2025demogen}, we synthesize trajectories by editing the 3D point cloud in one-shot demonstration: objects are translated across grid cells (aligned with Sec.~\ref{appdB1} standards) and rotated (for orientation-sensitive tasks) to match our BiDemoSyn spatial diversity for fair comparison. Edited point clouds exhibit artifacts (\textit{e.g.}, misaligned edges, self-occluded regions) due to perspective distortions, degrading observation fidelity. For YOTO \cite{zhou2025you}, we reduce data scale to 1/10 of BiDemoSyn to accommodate its time-consuming physical replay. Even at this scale, YOTO requires 388/777/115/259/129/100 replays across all six tasks (\textit{e.g.}, \texttt{plugpen} needs 388 replays, taking about 11 hours), versus BiDemoSyn's 3-hour synthesis for 10$\times$ data. Note that the efficiency recorded here is the time consumed by continuous collection, including a lot of non-collection time for transition and error recovery. The spatial sampling of YOTO is relaxed to coarse left-right object placements, sacrificing grid uniformity.

\begin{figure*}[h]
	\begin{center}
           \includegraphics[width=1.0\linewidth]{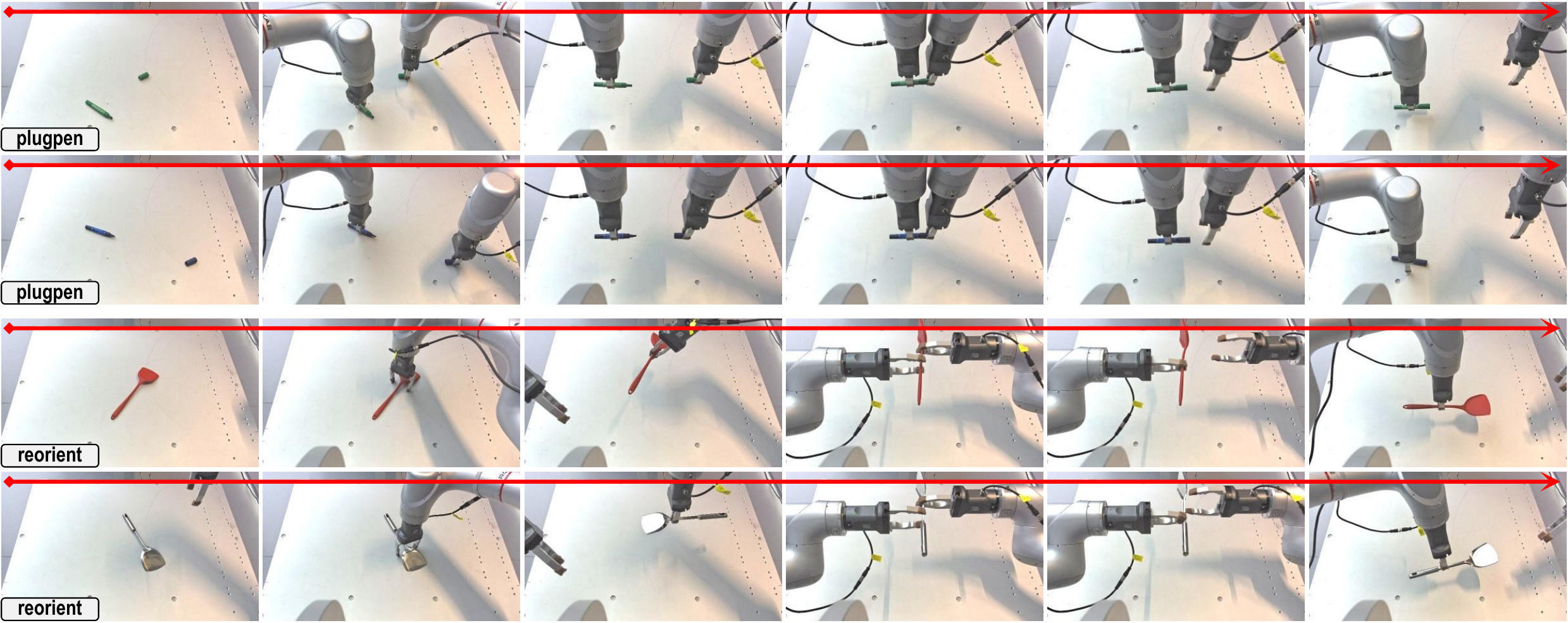}
	\vspace{-15pt}
	\caption{Qualitative real robot rollout samples for two tasks (\texttt{plugpen} and \texttt{reorient}) on the auxiliary dual-arm platform.}
           \label{rokaeVisStat}
	\vspace{-20pt}
	\end{center}
\end{figure*}

\subsection{Implementation of Imitation Policies}
Both DP3 \cite{ze2024dp3} and EquiBot \cite{yang2024equibot} (refer Fig.~\ref{networkDP}) ingest segmented 3D object point clouds but differ in network architectures. Specifically, DP3 employs specially designed multiple MLPs to encode point clouds into latent features, then directly regressing dual-arm keyposes. EquiBot enhances robustness via a SIM(3)-equivariant PointNet variant: point clouds undergo rotation/scale-invariant feature extraction before action prediction. This equips EquiBot with improved generalization to unseen object variations, achieving 5–12\% higher success rates than DP3 under identical training data and settings. Both policies operate in open-loop, aligning with synthesized demonstration formats. Below, we give a more detailed description of the visuomotor policy modification for the dual-arm action prediction.

\textbf{Spaces of Observation and Action}. We adopt a 13-dimensional proprioception vector and a 7-dimensional action space for each robot arm, respectively. The proprioception data for each arm consists of the following information: a 3-dimensional end-effector position, a 6-dimensional vector denoting end-effector orientation (represented by two columns of the end-effector rotation matrix), a 3-dimensional vector indicating the direction of gravity, and a scalar that represents the degree to which the gripper is opened. The action space for each arm consists of the following information: a 3-dimensional vector for the end-effector position offset, a 3-dimensional vector for the end-effector angular velocity in axis-angle format, and a scalar denoting the gripper action. For all bimanual tasks, the observation horizon is set to 1, so we only use the initial state observation of left arm as one of the network inputs. And the initial state of right arm is always fixed in each task. For the number of action steps (also the length of the predicted horizon), we simplify it and set the prediction length to the number of keyposes $K$, which can be extracted from the start and end actions of each Atomic Execution Primitives (AEPs) during task deconstruction.

\textbf{Network Architecture}. For DP3, it is a variant of Diffusion Policy \cite{chi2023diffusion} with a simpler point cloud encoder. It also designs a two-layer MLP to encode robot proprioceptive states before concatenating with the observation representation. For EquiBot, we use a SIM(3)-equivariant PointNet++ \cite{yang2024equivact} with 4 layers and hidden dimensionality 128 as the feature encoder. For the noise prediction network, we inherits hyperparameters from the original Diffusion Policy. Specifically, to optimize for inference speed  in all experiments, we use the DDIM scheduler \cite{song2021denoising} with 8 denoising steps, instead of the DDPM scheduler \cite{ho2020denoising} which performs up to 100 denoising steps.

\textbf{Sampling of Point Cloud}. As we all known, setting the number of points to sample in the point cloud observation is a key hyperparameter to consider when designing an architecture that takes point cloud inputs. In our experiments, we found out that using 1024 points is sufficient for all tasks and policies. In particular, we have tried increasing the number of point clouds to 2048 or more, but the evaluation improvement in each task is minimal, and this will also cause the storage occupied by the training observation data to be too large and the training time cost to increase. Therefore, reducing the number of points to 1024 can make training faster without hurting performance. And all our policy models can be trained on a GeForce RTX 3090 Ti with 24 GB of memory.

\textbf{Training and Evaluation}. When conducting the in-distribution (ID) and out-of-distribution (OOD) evaluations with full demonstrations, we train all methods for 500 and 1,000 epochs, respectively. Otherwise, when using the 1/10 training data (especially the YOTO), we train all methods of the ID and OOD evaluations for 2,000 and 4,000 epochs, respectively. For all experiments, the batch-size is always set to 64. We only evaluate the last one checkpoint saved at the end of training. For every evaluation in the real world, we run the policy in a randomly initialized placement of objects, and record the average success rate achieved by the policy.

In addition, we have adopted the object-centric point cloud input. At inference time, we also need to preprocess the binocular RGB observations to obtain the point cloud of manipulated objects. This core design relies on the still rapidly developing capabilities of vision foundation models (VFMs). Here we leverage SOTA open vocabulary detection method Florence-2 \cite{xiao2024florence} and segmentation method SAM2 \cite{ravi2024sam} to automatically extract object masks and then filter out corresponding point clouds. Despite this, occasionally we may fail to segment desired objects accurately, and in these special cases we will manually correct the masks. These cases are not counted as failed evaluation trails due to not involving significant elements of bimanual robot manipulation. Because we believe that the next generation of VFMs can alleviate these problems, or we can directly address them through domain adaptation, test-time adaptation, or adjusting input prompts.

\section{More Results Visualization and Analysis}\label{appdD}

\subsection{More Examples from Real Rollouts}
Fig.~\ref{moreVis} supplements more qualitative results with additional real-robot execution visualizations, highlighting nuanced state transitions and task-specific challenges. These visualizations underscore BiDemoSyn's ability to handle real-world complexities (such as mechanical tolerances and imperfect perception), while also exposing limitations in dynamic force modulation (\textit{e.g.}, over-pressing bottles or lids). Please refer our \textbf{Supplemental Videos} which provide frame-by-frame analyses of these executions, further dissecting critical phase and offering insights into policy decision-making under uncertainty.

\subsection{Two More Complex Long-Horizon Tasks}
To further evaluate the scalability of BiDemoSyn beyond the introduced six bimanual tasks, we add two new long-horizon tasks that require sequential manipulation steps and accumulate more physical error throughout the execution:

\begin{itemize}
\item \texttt{reorient+unscrew}: The robot must first upright a lying down bottle (reorient phase), then perform a dual-arm unscrewing motion to remove the cap (unscrew phase). This task stresses both precise estimation of bottle orientation and robust reuse of multi-step, contact-rich interaction patterns.
\vspace{-2pt}
\item \texttt{unscrew+pouring}: The fisrt phaset is the original unscrew task. After unscrewing the bottle cap, the robot must lift a mug cup, perform dual-arm coordination to tilt the bottle, and pour water into the cup. This composition integrates two high-precision subtasks and requires stable arm coordination across multiple transitions.
\end{itemize}

To utilize BiDemoSyn, these two newly added tasks were synthesized following the same three-stage pipeline described in the main text. We briefly summarize the configurations: for \texttt{reorient+unscrew}, we collected initial observations for 4 bottle instances, each placed at 7 orientations across 36 grid cells, yielding \textbf{1008 synthesized demonstrations}; for \texttt{unscrew+pouring}, we collected initial observations for 2 bottle and 2 cup instances, each placed on 18 grid cells, yielding \textbf{1296 synthesized demonstrations}. Importantly, both tasks introduce \textit{longer temporal horizons, sequential coordination, and increased compounding error}, thus providing a stronger stress test than the original six tasks.

Then, using the synthesized datasets, we trained visuomotor policies for each long-horizon task based on the adapted EquiBot architecture. Across both tasks, the trained policies exhibited high robustness to spatial variation (different placements and bottle orientations), reliable execution of multi-step dual-arm coordination, and consistent task success under mild external perturbations introduced during pre-grasp interpolation. In Fig.~\ref{newTasks}, we present some qualitative results with additional real-robot execution visualizations of these two new tasks. And in Fig.~\ref{dynaVis}, we provide illustrative examples of continuous dynamic interference scenarios among tasks \texttt{pouring} and \texttt{reorient+unscrew}. Full dynamic demonstrations and rollouts of all eight tasks (previous 6 tasks and the new 2 tasks) are included in the \textbf{Supplemental Videos}. These results further validate BiDemoSyn's ability to rapidly scale to new, long-horizon tasks with minimal additional engineering, and reinforce its suitability as a practical framework for synthesizing large-scale, real-world training data.

\begin{figure*}[h]
	\begin{center}
           \includegraphics[width=1.0\linewidth]{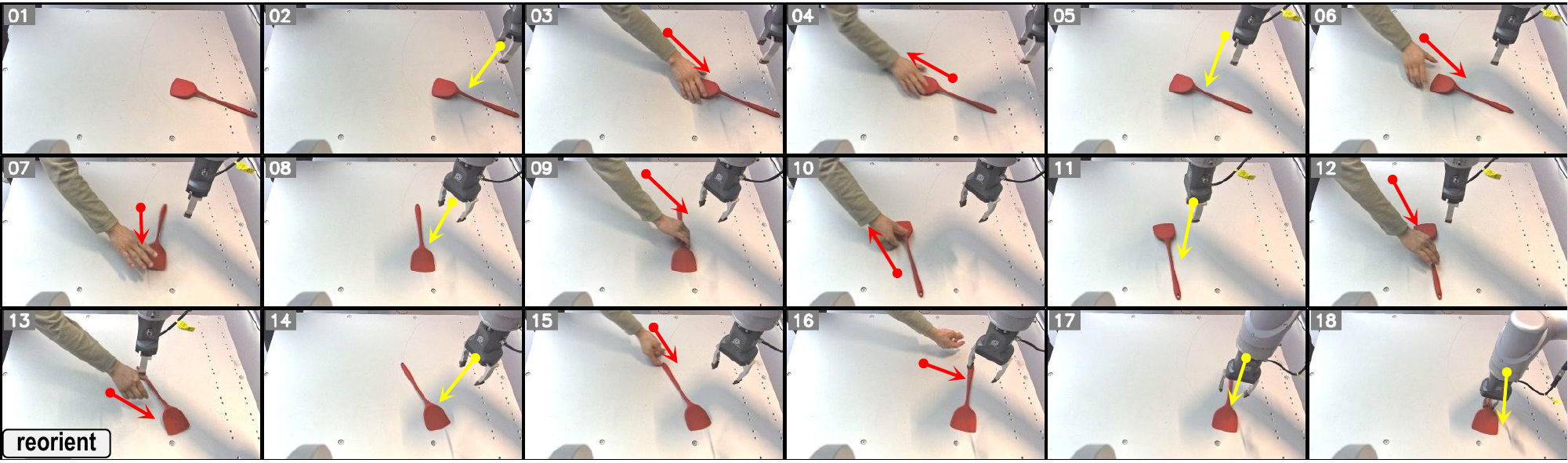}
	\vspace{-15pt}
	\caption{Qualitative examples of applying dynamic interferences in pre-grasping for the task \texttt{reorient} on the auxiliary dual-arm platform, where the object is manually disturbed up to five times. The red arrow indicates the direction of the manually moved object (interfering). The yellow arrow indicates the movement direction of the right robotic arm (chasing).}
           \label{rokaeVisDyna}
	\vspace{-20pt}
	\end{center}
\end{figure*}

\subsection{Cross-Embodiment Transferability Testing}\label{appdDrokae}
Beyond training and evaluating policies on the primary dual-arm platform (Fig.~\ref{platform} top), we further investigate the cross-embodiment transferability of policies trained with BiDemoSyn. Specifically, we deploy the learned visuomotor policies on a distinct humanoid-style dual-arm robot with different kinematic structures, arm mounting configurations, and workspace geometry (Fig.~\ref{platform} bottom). Importantly, no additional demonstrations, fine-tuning, or embodiment-specific retraining are introduced.

We evaluate two representative bimanual tasks—\texttt{plugpen} and \texttt{reorient}—which respectively emphasize precise dual-arm alignment and large-range end-effector rotation. As illustrated in Fig.~\ref{rokaeVisStat}, policies trained on the original platform can be directly executed on the new robot after a simple coordinate transformation of the predicted 6-DoF end-effector poses, without requiring explicit calibration of dynamic or contact parameters. Across extensive trials, both tasks achieve stable and consistent execution performance, comparable to those reported on the source platform. To further assess robustness, we introduce external perturbations during the pre-grasp stage (similar to Fig.~\ref{dynaVis}), including manual displacement and rotation of target objects. As shown in Fig.~\ref{rokaeVisDyna}, the policy can reliably recover from such disturbances and successfully complete the task, demonstrating that the transferred policy preserves not only task semantics but also robustness to moderate environmental variations. These qualitative results highlight a key advantage of BiDemoSyn: by adopting object-centric observations and embodiment-agnostic action representations, the resulting policies exhibit strong zero-shot transferability across heterogeneous robot embodiments, suggesting promising scalability to broader robotic platforms. More rollouts on this humanoid platform are in our \textbf{Supplemental Videos}. 

\begin{figure}[t]
	\begin{center}
           \includegraphics[width=\linewidth]{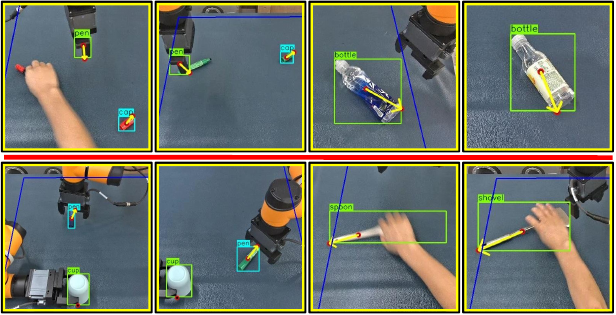}
	\vspace{-15pt}
	\caption{ \textit{(Top)} Failure cases in the task \texttt{plugpen} that wrongly detected the marker body, and the task \texttt{reorient+unscrew} that incorrectly estimated the lying bottle's orientation. \textit{(Bottom)} Failure cases in the task \texttt{inserting} where the robotic arm's end effector occasionally obscures the marker pen, and the task \texttt{reorient} where human interference causes the hand to partially obscure the manipulated spoon/shovel. } 
           \label{failedCases}
	\vspace{-25pt}
	\end{center}
\end{figure}

\begin{figure*}[t]
	\begin{center}
           \includegraphics[width=1.0\linewidth]{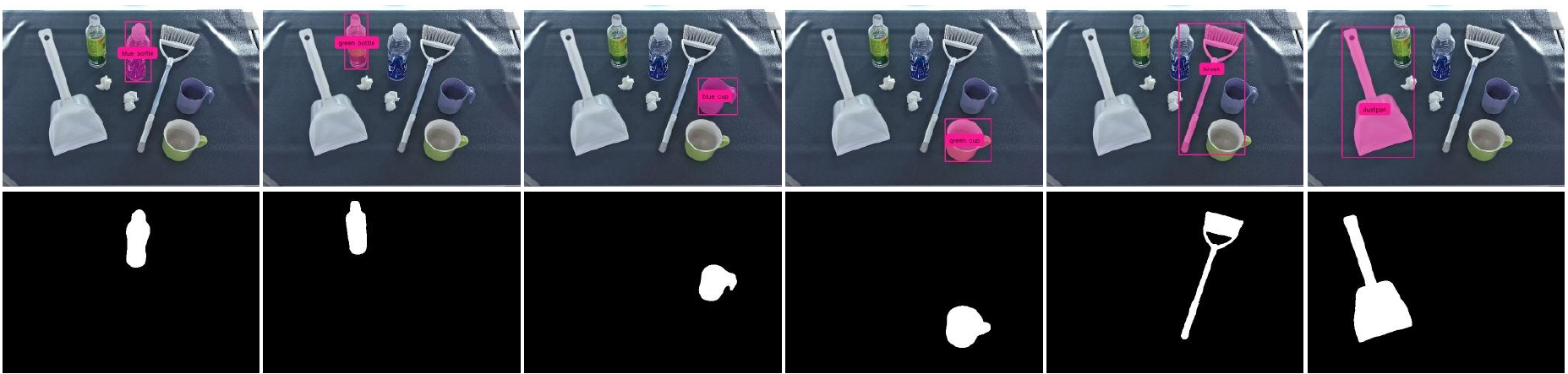}
	\vspace{-15pt}
	\caption{\textbf{Top}: Example cluttered tabletop scenes used to evaluate the perception front-end of BiDemoSyn. Each scene includes the target objects for the \texttt{pouring} task—two bottles (text prompts are \textit{blue bottle} and \textit{green bottle}) and two mugs (text prompts \textit{blue cup} and \textit{green cup})—surrounded by visually distracting household items. Two additional tool-like distractors (\textit{brush}, \textit{dustpan}) are intentionally introduced to support future extensions such as dual-arm \texttt{sweeping} tasks. \textbf{Bottom}: High-quality 2D object masks produced by Florence-2 + SAM2 for all queried categories. }
           \label{clutterSeg}
	\vspace{-20pt}
	\end{center}
\end{figure*}

\subsection{Statistics and Analysis of Failed Cases}\label{appdDfailure}
While policies trained with BiDemoSyn achieve high success rates, failure analysis reveals systematic challenges. Note that although the policy we trained is executed end-to-end (it establishes an implicit mapping from observations to action outputs), we can still align the analysis from the stage where its failure cases are located to find the core cause of the error. Finally, using the experimental results of EquiBot under out-of-distribution evaluations, we categorize failures into five types according to the task execution logic (mainly from the design module of BiDemoSyn):

\textbf{(1) Visual Perception Errors (15\%)}: Inaccurate object segmentation (\textit{e.g.}, mistaking the tabletop as part of the marker pen) leads to misaligned grasps.

\textbf{(2) Orientation Estimation Errors (32\%)}: Incorrect object pose estimation (\textit{e.g.}, misjudging a spoon's concave direction) causes malformed trajectories (\textit{e.g.}, flipping to the wrong side).

\textbf{(3) Alignment Failures (18\%)}: Vision-guided initial frame misalignment (\textit{e.g.}, bottle nozzle offset by $>$1cm) propagates errors to subsequent steps (\textit{e.g.}, spilling during \texttt{pouring}).

\textbf{(4) Initial Grasp Failures (28\%)}: Collisions during pre-grasp motions (\textit{e.g.}, gripper nudging objects) or unstable grasps (\textit{e.g.}, slippage in \texttt{reorient}) prevent task initiation.

\textbf{(5) Error Propagation (7\%)}: Cumulative errors from earlier stages compound into irreversible failures (\textit{e.g.}, slightly misaligned pluging between pen body and pen cap in \texttt{plugpen}).

As can be seen, the orientation estimation and initial grasp failures dominate, reflecting two core challenges: (1) current pose estimators struggle with symmetric or textureless objects (\textit{e.g.}, metal spoon or shovel), and (2) gripper-centric path planning lacks fine-grained contact modeling (\textit{e.g.}, avoiding pre-touch collisions for irregular shapes). Addressing these requires advances in category-agnostic pose estimation and short-horizon contact optimization, which are critical directions for our future work. Some visual failure examples are shown in Fig.~\ref{failedCases}, which provides a foundation for a full understanding of the BiDemoSyn system and its robust adaptation to allow for dynamic pre-grasping. More dynamic rollouts can be found in our \textbf{Supplementary Videos}.

\subsection{VFM-based Object Anchoring in Clutter}
To evaluate the applicability of BiDemoSyn under more realistic and visually complex environments, we additionally examine its object-anchoring capability when target objects are placed in cluttered scenes with distractors and partial occlusions. Although the main experiments intentionally isolate environmental variables to study the core challenge of real-world demonstration synthesis, BiDemoSyn is fully compatible with more complex perception conditions due to its reliance on VFMs for object segmentation.

We utilize Florence-2 \cite{xiao2024florence} and SAM2 \cite{ravi2024sam} to detect and segment task-relevant items using either text queries or exemplar-based prompts. Fig.~\ref{clutterSeg} illustrates representative examples where both bottle and mug are correctly segmented out of surrounding tools, utensils, packaging, and nonrigid paper balls. In cluttered tabletop scenes, the VFM consistently produces high-quality masks, even when distractors share similar color similarity and semantic ambiguity. Qualitatively, the predicted masks remain stable across moderate occlusions, enabling the system to isolate the target object with minimal ambiguity and providing reliable anchors for subsequent pose inference and variant-block adaptation in BiDemoSyn. 

Because BiDemoSyn's variant-block adaptation depends only on reliable 2D masks along with centroid- and principal-axis–based orientation inference, accurate object anchoring is the only requirement for extending the system to cluttered environments. Our experiments show that the VFM-based perception pipeline provides robust segmentation to support pose-aligned trajectory modulation and collision-aware grasp adjustments, even when the background is visually nonuniform. However, while VFMs significantly enhance robustness to visual complexity, extreme occlusions or heavy inter-object entanglement may still degrade performance. Extending BiDemoSyn to handle these cases will require multi-view sensing, active view planning, or integrating VLM-driven object-query disambiguation. Nevertheless, these initial results demonstrate that \textit{VFM-based anchoring provides a practical and reliable foundation for deploying BiDemoSyn beyond clean tabletop settings}, thereby enabling future extensions to household, warehouse, and mobile manipulation scenarios.

\section{Summary, Reflection and Outlook}\label{appdE}


This work introduces BiDemoSyn, a real-world demonstration synthesis framework that addresses a central bottleneck in bimanual imitation learning: how to scale high-quality, physically grounded data collection without reliance on simulation or extensive teleoperation. By decomposing a single kinesthetic demonstration into invariant coordination structures and object-dependent adaptations, and by coupling vision-guided alignment with lightweight trajectory optimization, BiDemoSyn enables the rapid generation of diverse, feasible bimanual demonstrations directly in the physical world. Extensive experiments across six contact-rich tasks demonstrate its efficiency, scalability, and strong generalization to novel object poses and geometries.

Beyond the original one-shot setting, we further show that BiDemoSyn naturally extends to \textit{few-shot demonstration synthesis}, where additional demonstrations primarily enhance out-of-distribution generalization without altering the core pipeline. Moreover, we demonstrate \textit{zero-shot cross-embodiment policy deployment}, where policies trained on BiDemoSyn data transfer effectively to a distinct humanoid-style dual-arm robot, preserving high success rates across representative tasks. These results highlight the importance of object-centric observations and embodiment-agnostic action representations, and suggest that BiDemoSyn supports not only scalable data generation, but also robust policy reuse across hardware platforms.

Despite these advances, several limitations remain. BiDemoSyn currently assumes quasi-static task execution and rigid or semi-rigid objects, making highly dynamic interactions and deformable object manipulation challenging. Failure cases are primarily attributed to perception noise—particularly orientation estimation for small or symmetric objects—and occasional contact inaccuracies during grasping and insertion. Addressing these issues will likely require tighter perception–control coupling, category-agnostic orientation reasoning, and more expressive contact-aware planners.

Looking forward, we see multiple promising directions. Future work will explore extending BiDemoSyn to deformable and articulated objects, incorporating dynamic and multi-view perception, and integrating closed-loop feedback into the synthesis pipeline. More broadly, we envision BiDemoSyn as a foundation for hybrid data-centric learning paradigms, combining one-shot, few-shot, and synthesized demonstrations to balance data efficiency with maximal generalization. By lowering the barrier to acquiring large-scale, real-world bimanual data, this work aims to accelerate progress toward practical, general-purpose robotic manipulation.


\end{document}